\documentclass[10pt,journal,compsoc,twoside]{IEEEtran}

\usepackage[nocompress]{cite}
\usepackage{graphicx}
\usepackage{amsmath}
\usepackage{amssymb}
\usepackage{amsfonts}
\usepackage{commath}
\usepackage{bm}
\usepackage{pifont}
\usepackage{subcaption}
\usepackage{float}
\usepackage{hyperref}
\hypersetup{colorlinks,allcolors=black}
\usepackage{tabularx}
\usepackage{xcolor}
\usepackage{booktabs}
\usepackage{adjustbox}

\usepackage{pgfplots}
\pgfplotsset{width=10cm,compat=1.9}
\usetikzlibrary{shapes.geometric,decorations.pathreplacing,shapes.arrows,angles,calc,quotes}
\usepackage{pgfplotstable}

\renewcommand{\vec}[1]{\boldsymbol{#1}}
\DeclareMathOperator{\loss}{\mathcal{L}}
\DeclareMathOperator{\E}{\mathbb{E}}
\DeclareMathOperator{\reals}{\mathbb{R}}

\newcount\myloopcounter
\newcommand{\nstars}[1][4]{%
  \myloopcounter0% initialize the loop counter
  \loop\ifnum\myloopcounter < 5
  \ifthenelse{\myloopcounter < #1}{
    \textcolor{black}{\star}
  }{
    \textcolor{black!22}{\star}
  }
  \advance\myloopcounter by 1 % 
  \repeat % start again
}

\begin{document}

% Colours
\definecolor{myblue}{HTML}{2d70b3}
\definecolor{myorange}{HTML}{fa7e19}
\definecolor{mygreen}{HTML}{388c46}
\definecolor{myred}{HTML}{c74440}
\definecolor{mypurple}{HTML}{6042a6}
\definecolor{mygrey}{HTML}{808080}

\title{Deep Generative Modelling: A Comparative Review of VAEs, GANs, Normalizing Flows, Energy-Based and Autoregressive Models}

\author{Sam~Bond-Taylor,
        Adam~Leach,
        Yang~Long,
        Chris~G.~Willcocks% <-this % stops a space
\IEEEcompsocitemizethanks{\IEEEcompsocthanksitem The authors are with the Department of Computer Science, Durham
University, Durham, DH1 3LE, United Kingdom. This work was supported by MRC Innovation Fellowship, ref MR/S003916/1.}

%The authors are with the Department of Computer Science, Durham University, Durham, DH1 3LE, UK. Email: \{samuel.e.bond-taylor, adam.leach, yang.long, christopher.g.willcocks\}@durham.ac.uk}

% \thanks{Manuscript received 14 Apr. 21; revised 5 July 21; accepted 14 Sept. 21\\Date of publication 0.0000; date of current version 0.0000.\\This work was supported by MRC Innovation Fellowship, ref MR/S003916/1.\\Recommded for accepance by O. Winther.\\Digital Object Identifier no. 10.1109/TPAMI.2021.3116668}

%This work has been submitted to the IEEE for possible publication. Copyright may be transferred without notice, after which this version may no longer be accessible.}
%of Electrical and Computer Engineering, Georgia Institute of Technology, Atlanta,
%GA, 30332.\protect\\
%E-mail: see http://www.michaelshell.org/contact.html
%\IEEEcompsocthanksitem J. Doe and J. Doe are with Anonymous University.}% <-this % stops an unwanted space
}

\markboth{IEEE TRANSACTIONS ON PATTERN ANALYSIS AND MACHINE INTELLIGENCE}%
{Bond-Taylor et al.: Deep Generative Modelling: A Comparative Review}

% Abstract and Keywords
\IEEEtitleabstractindextext{%
\begin{abstract}
Deep generative models are a class of techniques that train deep neural networks to model the distribution of training samples. Research has fragmented into various interconnected approaches, each of which make trade-offs including run-time, diversity, and architectural restrictions. In particular, this compendium covers energy-based models, variational autoencoders, generative adversarial networks, autoregressive models, normalizing flows, in addition to numerous hybrid approaches. These techniques are compared and contrasted, explaining the premises behind each and how they are interrelated, while reviewing current state-of-the-art advances and implementations.
\end{abstract}

}

\maketitle

\IEEEdisplaynontitleabstractindextext
\IEEEpeerreviewmaketitle

\IEEEraisesectionheading{\section{Introduction}\label{sec:introduction}}

\IEEEPARstart{G}{enerative} modelling using neural networks has its origins in the 1980s with aims to learn about data with no supervision, potentially providing benefits for standard classification tasks; collecting training data for unsupervised learning is naturally much lower effort and cheaper than collecting labelled data but there is considerable information still available making it clear that generative models can be beneficial for a wide variety of applications. 

Beyond this, generative modelling has numerous direct applications including image synthesis: super-resolution, text-to-image and image-to-image conversion, inpainting, attribute manipulation, pose estimation; video: synthesis and retargeting; audio: speech and music synthesis; text: summarisation and translation; reinforcement learning; computer graphics: rendering, texture generation, character movement, liquid simulation; medical: drug synthesis, modality conversion; and out-of-distribution detection.%; and feature generation.

The central idea of generative modelling stems around training a generative model whose samples $\tilde{\vec{x}} \sim p_\theta(\tilde{\vec{x}})$ come from the same distribution as the training data distribution, $\vec{x} \sim p_d(\vec{x})$. Early neural generative models, energy-based models achieved this by defining an energy function on data points proportional to likelihood, however, these struggled to scale to complex high dimensional data such as natural images, and require Markov Chain Monte Carlo (MCMC) sampling during both training and inference, a slow iterative process. In recent years there has been renewed interest in generative models driven by the advent of large freely available datasets as well as advances in both general deep learning architectures and generative models, breaking new ground in terms of visual fidelity and sampling speed. In many cases, this has been achieved using latent variables $\vec{z}$ which are easy to sample from and/or calculate the density of, instead learning $p(\vec{x}, \vec{z})$; this requires marginalisation over the unobserved latent variables, however in general, this is intractable. Generative models therefore typically make trade-offs in execution time, architecture, or optimise proxy functions. Choosing what to optimise for has implications for sample quality, with direct likelihood optimisation often leading to worse sample quality than alternatives.

Interrelated with generative models is the field of self-supervised learning where the focus is on learning good intermediate representations that can be used for downstream tasks without supervision \cite{jing2020self}. As such, generative models can in general also be considered self-supervised, however, not all self-supervised models are generative models. Types of self-supervised objectives include auxiliary classification losses such as predicting the rotation of inputs, masked losses where the model must predict the true value of some inputs which have been masked out, and contrastive losses which learn an embedding space where similar data points are close and different points are far apart.

\begin{table*}[t]
  \caption{Comparison between deep generative models in terms of training and test speed, parameter efficiency, sample quality, sample diversity, and ability to scale to high resolution data. Quantitative evaluation is reported on the CIFAR-10 dataset \cite{krizhevsky2009learning} in terms of Fr\'echet Inception Distance (FID) and negative log-likelihood (NLL) in bits-per-dimension (BPD).}
  \label{tab:methods-compared}
  \centering
  \begin{tabular}{>{\raggedright}p{47mm}>{\centering\arraybackslash}p{10mm}>{\centering\arraybackslash}p{10mm}>{\centering\arraybackslash}p{10mm}>{\centering\arraybackslash}p{10mm}>{\centering\arraybackslash}p{12.4mm}>{\centering\arraybackslash}p{10mm}>{\centering\arraybackslash}p{10mm}>{\centering\arraybackslash}p{10mm}}
    \toprule
    Method         & Train Speed & Sample Speed & Num. Params. & Resolution Scaling & Free-form Jacobian & Exact Density & FID & NLL (in BPD) \\
    \midrule
    Generative Adversarial Networks & & & & & \\
    \quad DCGAN \cite{Radford2016UnsupervisedRepresentationLearning}                                        & $\nstars[5]$ & $\nstars[5]$ & $\nstars[5]$ & $\nstars[3]$ & \textcolor{black}{\ding{51}}  & \textcolor{black}{\ding{55}} & 37.11  & -              \\
    \quad ProGAN \cite{Karras2018ProgressiveGrowingGANsa}                                                   & $\nstars[4]$ & $\nstars[5]$ & $\nstars[4]$ & $\nstars[5]$ & \textcolor{black}{\ding{51}}  & \textcolor{black}{\ding{55}} & 15.52  & -              \\
    \quad BigGAN \cite{Brock2019LargeScaleGAN}                                                              & $\nstars[4]$ & $\nstars[5]$ & $\nstars[5]$ & $\nstars[4]$ & \textcolor{black}{\ding{51}}  & \textcolor{black}{\ding{55}} & 14.73  & -              \\
    \quad StyleGAN2 + ADA \cite{karras2020training}                                                         & $\nstars[1]$ & $\nstars[5]$ & $\nstars[3]$ & $\nstars[5]$ & \textcolor{black}{\ding{51}}  & \textcolor{black}{\ding{55}} & 2.42   & -              \\
    \midrule
    Energy Based Models & & & & & \\
    \quad IGEBM \cite{Du2019ImplicitGenerationGeneralization}                                               & $\nstars[3]$ & $\nstars[2]$ & $\nstars[5]$ & $\nstars[3]$ & \textcolor{black}{\ding{51}}  & \textcolor{black}{\ding{55}} & 37.9   & -              \\
    \quad Denoising Diffusion \cite{Ho2020DenoisingDiffusionProbabilistic}                                  & $\nstars[5]$ & $\nstars[2]$ & $\nstars[3]$ & $\nstars[4]$ & \textcolor{black}{\ding{51}}   & \textcolor{mygrey}{(\ding{51})} & 3.17   & $\leq 3.75$    \\
    \quad DDPM++ Continuous \cite{song2021scorebased}                                                       & $\nstars[5]$ & $\nstars[2]$ & $\nstars[3]$ & $\nstars[4]$ & \textcolor{black}{\ding{51}}   & \textcolor{mygrey}{(\ding{51})} & 2.20   & -           \\
    \quad Flow Contrastive (EBM) \cite{gao2020flow}                                                               & $\nstars[1]$ & $\nstars[2]$ & $\nstars[2]$ & $\nstars[2]$ & \textcolor{black}{\ding{51}}   & \textcolor{black}{\ding{55}} & 37.30  & $\approx 3.27$ \\
    \quad VAEBM \cite{xiao2021vaebm}                                                                        & $\nstars[2]$ & $\nstars[4]$ & $\nstars[1]$ & $\nstars[4]$ & \textcolor{black}{\ding{51}}  & \textcolor{black}{\ding{55}} & 12.19  & -              \\
    \midrule
    Variational Autoencoders & \\
    \quad Convolutional VAE \cite{Kingma2014AutoEncodingVariationalBayes}                                   & $\nstars[5]$ & $\nstars[5]$ & $\nstars[5]$ & $\nstars[2]$ & \textcolor{black}{\ding{51}}  & \textcolor{mygrey}{(\ding{51})} & 106.37 & $\leq 4.54$    \\
    \quad Variational Lossy AE \cite{Chen2017VariationalLossyAutoencoder}                                  & $\nstars[3]$ & $\nstars[1]$ & $\nstars[1]$ & $\nstars[2]$ & \textcolor{black}{\ding{55}}    & \textcolor{mygrey}{(\ding{51})} & -      & $\leq 2.95$    \\
    \quad VQ-VAE \cite{VanDenOord2017NeuralDiscreteRepresentation,Razavi2019GeneratingDiverseHighFidelity} & $\nstars[2]$ & $\nstars[3]$ & $\nstars[1]$ & $\nstars[5]$ & \textcolor{black}{\ding{55}}    & \textcolor{mygrey}{(\ding{51})} & -      & $\leq 4.67$    \\
    \quad VD-VAE \cite{child2021very}                                                                       & $\nstars[1]$ & $\nstars[5]$ & $\nstars[3]$ & $\nstars[5]$ & \textcolor{black}{\ding{51}} & \textcolor{mygrey}{(\ding{51})} & -      & $\leq 2.87$    \\
    \midrule
    Autoregressive Models & & & & & \\
    \quad PixelRNN \cite{VanDenOord2016PixelRecurrentNeural}                                                & $\nstars[2]$ & $\nstars[1]$ & $\nstars[1]$ & $\nstars[3]$ & \textcolor{black}{\ding{55}}    & \textcolor{black}{\ding{51}} & -      & 3.00           \\
    \quad Gated PixelCNN \cite{VanDenOord2016ConditionalImageGeneration}                                    & $\nstars[3]$ & $\nstars[1]$ & $\nstars[1]$ & $\nstars[3]$ & \textcolor{black}{\ding{55}}    & \textcolor{black}{\ding{51}} & 65.93  & 3.03           \\
    \quad PixelIQN \cite{Ostrovski2018AutoregressiveQuantileNetworks}                                       & $\nstars[3]$ & $\nstars[1]$ & $\nstars[1]$ & $\nstars[3]$ & \textcolor{black}{\ding{55}}    & \textcolor{black}{\ding{51}} & 49.46  & -              \\
    \quad Sparse Trans. + DistAug \cite{Child2019GeneratingLongSequences,jun2020distribution}               & $\nstars[4]$ & $\nstars[1]$ & $\nstars[3]$ & $\nstars[3]$ & \textcolor{black}{\ding{55}}    & \textcolor{black}{\ding{51}} & 14.74  & 2.66           \\
    \midrule
    Normalizing Flows & & & & & \\
    \quad RealNVP \cite{Dinh2016DensityEstimationUsing}                                                     & $\nstars[2]$ & $\nstars[5]$ & $\nstars[5]$ & $\nstars[3]$ & \textcolor{black}{\ding{55}}    & \textcolor{black}{\ding{51}} & -       & 3.49          \\
    \quad GLOW \cite{Kingma2018GlowGenerativeFlow}                                                          & $\nstars[1]$ & $\nstars[5]$ & $\nstars[2]$ & $\nstars[4]$ & \textcolor{black}{\ding{55}}    & \textcolor{black}{\ding{51}} & 45.99   & 3.35          \\
    \quad FFJORD \cite{Grathwohl2018FFJORDFreeformContinuous}                                               & $\nstars[1]$ & $\nstars[3]$ & $\nstars[5]$ & $\nstars[2]$ & \textcolor{black}{\ding{51}}  & \textcolor{mygrey}{(\ding{51})} & -       & 3.40          \\
    \quad Residual Flow \cite{chen2019residual}                                                             & $\nstars[2]$ & $\nstars[4]$ & $\nstars[4]$ & $\nstars[4]$ & \textcolor{black}{\ding{51}}  & \textcolor{mygrey}{(\ding{51})} & 46.37   & 3.28          \\
    \bottomrule
  \end{tabular}
\end{table*}

There exists a variety of survey papers focusing on particular generative models such as normalizing flows \cite{kobyzev2020normalizing, Papamakarios2019NormalizingFlowsProbabilistic}, generative adversarial networks \cite{gui2020review, yi2019generative}, and energy-based models \cite{song2021how}, however, naturally these dive into the intricacies of their respective method rather than comparing with other methods; additionally, some focus on applications rather than theory. While there exists a recent survey on generative models as a whole \cite{oussidi2018deep}, it is less broad, diving deeply into a few specific implementations.

This survey provides a comprehensive overview of generative modelling trends, introducing new readers to the field, comparing and contrasting so as to explain the modelling decisions behind each respective technique. Additionally, advances old and new are discussed in order to bring the reader up to date with current research. A specific focus on image models is taken reflecting the predominance in literature, however, concepts are often relevant across modalities. In particular, this survey covers \hyperref[sec:ebm]{energy-based models}, unnormalised density models, \hyperref[sec:vae]{variational autoencoders}, variational approximation of a latent-based model's posterior, \hyperref[sec:gan]{generative adversarial networks}, two models set in a mini-max game, \hyperref[sec:autoregressive]{autoregressive models}, model data decomposed as a product of conditional probabilities, and \hyperref[sec:normalizing-flows]{normalizing flows}, exact likelihood models using invertible transformations. This breakdown is defined to closely match the typical divisions within research, however, numerous hybrid approaches exist that blur these lines, these are discussed in the most relevant section or both where suitable.

\begin{table}[t]
    \caption{Rules for the star ratings in Table \ref{tab:methods-compared}.}
    \label{tab:star-system}
    \centering
    \begin{tabular}{p{10mm}ccccc}
        \toprule
                   & 1 Star    & 2 Stars      & 3 Stars      & 4 Stars        & 5 Stars \\ \midrule
        Training   & $>$5 days & $\leq$5 days & $\leq$2 days & $\leq$1 days   & $\leq\hspace{-0.4em}\frac{1}{2}$ day \\
        Sampling   & AR        & MCMC         & Middle       & $\leq$20 steps & 1 step  \\
        Params     & $>$120M   & $\leq$120M   & $\leq$60M    & $\leq$30M      & $\leq$10M \\
        Resolution & $<$32     & 32           & 64 or 128    & 256 or 512     & $\geq$1024  \\ \bottomrule
    \end{tabular}
    \vspace*{-1em}
\end{table}

For a brief insight into the differences between architectures, we provide Table \ref{tab:methods-compared} which contrasts a diverse array of techniques. For the column ``Exact Density'', \textcolor{black}{\ding{51}} represents tractable densities, \textcolor{mygrey}{(\ding{51})} approximate densities, and \textcolor{black}{\ding{55}} intractable densities. On a number properties assessed we use a star system to allow easy comparisons, with rules defined in Table \ref{tab:star-system} based on CIFAR-10. In particular, we acknowledge that ranking measures such as training speed in days can be considered anecdotal since it is dependent on the year and compute available. Nevertheless, this allows a comparison based on properties such as stability and convergence rates which cannot be easily judged, for instance, by simply looking at number of function evaluations per iteration.

%%%%%%%%%%%%%%%%%%%%%%%%%%%%%%%%%%%%%%%%%%%%
%%%%%%%%%%%%%%%%%%%%%%%%%%%%%%%%%%%%%%%%%%%%
%%%%%%%%%%% ENERGY-BASED MODELS %%%%%%%%%%%%
%%%%%%%%%%%%%%%%%%%%%%%%%%%%%%%%%%%%%%%%%%%%
%%%%%%%%%%%%%%%%%%%%%%%%%%%%%%%%%%%%%%%%%%%%

% Unnormalised probability models
\section{Energy-Based Models \label{sec:ebm}}
Energy-based models (EBMs) \cite{LeCun2006TutorialEnergyBasedLearning} are based on the observation that any probability density function $p(\vec{x})$ for $\vec{x} \in \reals^D$ can be expressed in terms of an energy function $E(\vec{x}) \colon \reals^D \to \reals$  which associates realistic points with low values and unrealistic points with high values
\begin{equation}\label{eqn:ebm-pdf}
    p(\vec{x}) = \frac{e^{-E(\vec{x})}}{\int_{\tilde{\vec{x}} \in \mathcal{X}} e^{-E(\tilde{\vec{x}})}}. 
\end{equation}
Modelling data in such a way offers a number of perks, namely the simplicity and stability associating with training a single model; utilising a shared set of features thereby minimising required parameters; and the lack of any prior assumptions eliminates related bottlenecks \cite{Du2019ImplicitGenerationGeneralization}. Despite these benefits, scaling to high dimensional data is difficult, however, recent advances have made substantial strides.

A key issue with EBMs is how to optimise them; since the denominator in Eqn. \ref{eqn:ebm-pdf} is intractable for most models, a popular proxy objective is contrastive divergence where energy values of data samples are `pushed' down, while samples from the energy distribution are `pushed' up. Formally, the gradient of the negative log-likelihood loss $\loss(\theta) = \E_{\vec{x}\sim p_d}[ -\ln p_\theta(\vec{x}) ]$ has been shown to approximately demonstrate the following property \cite{Carreira-Perpinan2005ContrastiveDivergenceLearning, Sutskever2010ConvergencePropertiesContrastive},
\begin{equation}\label{eqn:ebm-training}
    \nabla_\theta \mathcal{L} = \E_{\vec{x}^+ \sim p_d}[ \nabla_\theta E_\theta(\vec{x}^+) ] - \E_{\vec{x}^- \sim p_\theta}[ \nabla_\theta E_\theta(\vec{x}^-) ],
\end{equation}
where $\vec{x}^- \sim p_\theta$ is a sample from the EBM found through a Markov Chain Monte Carlo (MCMC) generating procedure.

%%%%%%%%%%%%%%%%%%%%%%%%%%%%%%%%%%%%%%%%%%%%
%%%%%%%%%%% Boltzmann Machines %%%%%%%%%%%%%
%%%%%%%%%%%%%%%%%%%%%%%%%%%%%%%%%%%%%%%%%%%%

\subsection{Early Energy-Based Models}
Before moving to recent advances, we start with some of the earliest neural generative models.

\subsubsection{Boltzmann Machines}
A Boltzmann machine \cite{Hinton1983OptimalPerceptualInference} is a fully connected undirected network of binary neurons (Fig. \ref{fig:boltzmann-machine}) that are turned on with probability determined by a weighted sum of their inputs i.e. for some state $s_i$, $p(s_i=1)=\sigma(\sum_j w_{i,j} s_j)$. The neurons can be divided into visible $\vec{v} \in \{0,1\}^D$ units, those which are set by inputs to the model, and hidden $\vec{h} \in \{0,1\}^P$ units, all other neurons. The energy of the state $\{\vec{v},\vec{h}\}$ is defined (without biases for succinctness) as
\begin{equation}
    E_\theta(\vec{v}, \vec{h}) =  -\frac{1}{2}\vec{v}^T\vec{L}\vec{v} -\frac{1}{2}\vec{h}^T\vec{J}\vec{h} -\frac{1}{2}\vec{v}^T\vec{W}\vec{h},
\end{equation}
where $\vec{W}$, $\vec{L}$, and $\vec{J}$ are symmetrical learned weight matrices. In order to train Boltzmann machines via contrastive divergence, equilibrium states are found via Gibbs sampling, however, this takes an exponential amount of time in the number of hidden units making scaling impractical.

\subsubsection{Restricted Boltzmann Machines}
\begin{figure}[t]
    \centering
    \begin{subfigure}{0.39\linewidth}
        \centering
        \begin{tikzpicture}[shorten >=1pt]
            \tikzstyle{cir}=[circle,fill=black!12,minimum size=17pt,inner sep=0pt]
            \tikzstyle{dia}=[diamond,fill=black!30,minimum size=19pt,inner sep=0pt]
            \foreach \name/\angle/\text in {P-1/234/v_1, P-5/-54/v_2}
                \node[cir,xshift=1cm,yshift=1cm] (\name) at (\angle:1cm) {$\text$};
            
            \foreach \name/\angle/\text in {P-2/162/h_3, P-3/90/h_2, P-4/18/h_1}
                \node[dia,xshift=1cm,yshift=1cm] (\name) at (\angle:1cm) {$\text$};
            
            \foreach \from/\to in {1/2,2/3,3/4,4/5,5/1,1/3,2/4,3/5,4/1,5/2}
            { \draw (P-\from) -- (P-\to); }
        \end{tikzpicture}
        \caption{Boltzmann machine.}
        \label{fig:boltzmann-machine}
    \end{subfigure}
    \begin{subfigure}{0.59\linewidth}
        \centering
        \vspace{0.8em}
        \begin{tikzpicture}[shorten >=1pt]
            \tikzstyle{cir}=[circle,fill=black!30,minimum size=17pt,inner sep=0pt]
            \tikzstyle{dia}=[diamond,fill=black!30,minimum size=19pt,inner sep=0pt]
            \tikzstyle{box}=[rounded corners=6pt,fill=black!25,minimum size=17pt,inner xsep=4pt, inner ysep=0pt]
            \tikzstyle{sqr}=[fill=black!25,minimum size=17pt, inner ysep=0pt, inner xsep=4pt]
            
            \node[cir,fill=black!12]  (RBM-1)    at (0,0) {$v_1$};
            \node[cir,fill=black!12]  (RBM-2)    at (1,0) {$v_2$};
            \node[cir,fill=black!12]  (RBM-3)    at (2,0) {$v_3$};
            \node[align=center] at (2.75,0) {$...$};
            \node[cir,fill=black!12]  (RBM-4)    at (3.5,0) {$v_n$};
            
            \node[dia]  (RBM-5)    at (0.5,1.3) {$h_1$};
            \node[dia]  (RBM-6)    at (1.5,1.3) {$h_2$};
            \node[align=center]    at (2.25,1.3) {$...$};
            \node[dia]  (RBM-7)    at (3,1.3) {$h_n$};
            
            \foreach \from/\to in {1/5,2/5,3/5,4/5,1/6,2/6,3/6,4/6,1/7,2/7,3/7,4/7}
            { \draw (RBM-\from) -- (RBM-\to); }
    \end{tikzpicture}
    \vspace{0.8em}
    \caption{Restricted Boltzmann machine.}
    \label{fig:restricted-boltzmann-machine}
    \end{subfigure}
    \caption{Restricted Boltzmann machines have restricted architectures to allow faster sampling than Boltzmann machines.}
    
\end{figure}
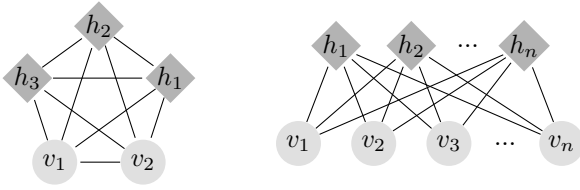

Many of the issues associated with Boltzmann machines can be overcome by restricting their connectivity. One approach, known as the restricted Boltzmann machine (RBM) \cite{Hinton2002TrainingProductsExperts} is to remove connections between units in the same group (Fig. \ref{fig:restricted-boltzmann-machine}), allowing exact calculation of hidden units. Although obtaining negative samples still requires Gibbs sampling, it can be parallelised and in practice a single step is sufficient if $\vec{v}$ is initially sampled from the dataset \cite{Hinton2002TrainingProductsExperts}.

By stacking RBMs, using features from lower down as inputs for the next layer, more powerful functions can be learned; these models are known as deep belief networks \cite{Hinton2006FastLearningAlgorithm}. Training an entire model at once is intractable so instead they are trained greedily layer by layer, composing densities thus improving the approximation of $p(\vec{v})$. 

%%%%%%%%%%%%%%%%%%%%%%%%%%%%%%%%%%%%%%%%%%%%
%%%%%%%%%%%%% RECENT ADVANCES %%%%%%%%%%%%%%
%%%%%%%%%%%%%%%%%%%%%%%%%%%%%%%%%%%%%%%%%%%%

\subsection{Deep EBMs via Contrastive Divergence}
To train more powerful architectures through contrastive divergence, one must be able to efficiently sample from $p_{\theta}$. Specifically, we would like to model high dimensional data using an energy function with a deep neural network, taking advantage of recent advances in discriminative models \cite{Zagoruyko2017WideResidualNetworks}. MCMC methods such as random walk and Gibbs sampling \cite{Hinton2006FastLearningAlgorithm}, when applied to high dimensional data, have long mixing times, making them impractical. 
A number of recent approaches \cite{Xie2016TheoryGenerativeConvNet, Du2019ImplicitGenerationGeneralization}
have advocated the use of stochastic gradient Langevin dynamics \cite{roberts1996exponential, Welling2011BayesianLearningStochastic} which permits sampling through the following iterative process,
\begin{equation}
    \vec{x}_0 \sim p_0(\vec{x}) \qquad \vec{x}_{i+1} = \vec{x}_i - \frac{\alpha}{2} \frac{\partial E_\theta(\vec{x}_i)}{\partial \vec{x}_i} + \vec{\epsilon}, \label{eqn:langevin-step}
\end{equation}
where $\vec{\epsilon} \sim \mathcal{N}(\vec{0},\alpha\vec{I})$, $p_0(\vec{x})$ is typically a uniform distribution over the input domain and $\alpha$ is the step size. As the number of updates $N \rightarrow \infty$ and $\alpha \rightarrow 0$, the distribution of samples converges to $p_{\theta}$ \cite{Welling2011BayesianLearningStochastic}; however, $\alpha$ and $\vec\epsilon$ are often tweaked independently to speed up training.

While Langevin MCMC is more practical than other approaches, sampling still requires a large number of steps. One solution is to use persistent contrastive divergence \cite{tieleman2008training, Du2019ImplicitGenerationGeneralization} where a replay buffer stores previously generated samples that are randomly reset to noise; this allows samples to be continually refined with a relatively small number of steps while maintaining diversity. Short-run MCMC \cite{Nijkamp2020AnatomyMCMCBasedMaximum} which samples using as few as 100 update steps from noise has also been used to train deep EBMs, however, since the number of steps is so small, samples are not truly from the correct probability density. Nevertheless, there are other advantages such as allowing image interpolation and reconstruction (since short-run MCMC does not mix) \cite{Nijkamp2019LearningNonConvergentNonPersistent}. Other approaches include initialising MCMC chains with data points \cite{Xie2016TheoryGenerativeConvNet} and samples from an implicit generative model \cite{xie2020cooperative}, as well as adversarially training an implicit generative model, mitigating mode collapse somewhat by maximising its entropy \cite{kumar2019maximum, grathwohl2021no, kim2016deep}. Improved/augmented MCMC samplers with neural networks can also improve the efficiency of sampling \cite{song2017anicemc, hoffman2019neutralizing, titsias2019gradientbased, li2020neural, grathwohl2021oops}.

One application of EBMs of this form comes by using standard classifier architectures, $f_\theta \colon \reals^D \to \reals^K$, which map data points to logits used by a softmax function to compute $p_\theta(y|\vec{x})$. By marginalising out $y$, these logits can be used to define an energy model that can be simultaneously trained as both a generative and classification model \cite{Grathwohl2019YourClassifierSecretly}, 
\begin{subequations}
\begin{equation}
    p_\theta(\vec{x}) = \sum_y p_\theta(\vec{x},y) = \frac{\sum_y \exp(f_\theta(\vec{x}[y]))}{Z(\theta)},
\end{equation}
\begin{equation}
    E_\theta(\vec{x}) = - \ln \sum_y \exp(f_\theta(\vec{x}[y])).
\end{equation}
\end{subequations}

\subsection{Score Matching and Denoising Diffusion}
Although Langevin MCMC has allowed EBMs to scale to high dimensional data, training times are still slow due to the need to sample from the model distribution, additionally, the finite nature of the sampling process means that samples can be arbitrarily far away from the model's distribution \cite{grathwohl2020learning}. An alternative approach is score matching \cite{Hyvarinen2005EstimationNonNormalizedStatistical} which is based on the idea of minimising the difference between the derivatives of the data and model's log-density functions; the score function is defined as $s(\vec{x}) = \nabla_{\vec{x}} \ln p(\vec{x})$ which does not depend on the intractable denominator and can therefore be applied to build an energy model \cite{Swersky2011AutoencodersScoreMatching} by minimising the Fisher divergence between $p_d$ and $p_\theta$,
\begin{equation}
    \mathcal{L} = \frac{1}{2}\E_{p_d(\vec{x})}[\text{ } \norm{s_\theta(\vec{x}) - s_d(\vec{x})}_2^2], \label{eqn:score-matching}
\end{equation}
however, the score function of data is usually not available. Various methods exist to estimate the score function including spectral approximation \cite{shi2018spectral}, sliced score matching \cite{song2020sliced}, finite difference score matching \cite{pang2020efficient}, and notably denoising score matching \cite{Vincent2011ConnectionScoreMatching} which allows the score to be approximated using corrupted data samples  $q(\tilde{\vec{x}}|\vec{x})$. In particular, when $q=\mathcal{N}(\tilde{\vec{x}} | \vec{x}, \sigma^2\bm{I})$, Eqn. \ref{eqn:score-matching} simplifies to
\begin{equation}
    \mathcal{L} = \frac{1}{2} \E_{p_d(\vec{x})}\E_{\tilde{\vec{x}}\sim\mathcal{N}(\vec{x},\sigma^2 \bm{I})} \bigg[ \norm{s_\theta(\tilde{\vec{x}}) + \frac{\tilde{\vec{x}} - \vec{x}}{\sigma^2} }^2_2 \bigg].
    \label{eqn:denoising-score-matching}
\end{equation}
That is, $s_\theta$ learns to estimate the noise thereby allowing it to be used as a generative model \cite{Song2019GenerativeModelingEstimating, saremi2018deep}. Since the Langevin update step uses $\nabla_{\vec{x}} \ln p(\vec{x})$ it is possible to sample from a score matching model using Langevin dynamics \cite{tzen2019theoretical}. This is only possible, however, when trained over a large variety of noise levels so that $\tilde{\vec{x}}$ covers the whole space. 

\subsubsection{Denoising Diffusion Probabilistic Models}
Closely related are diffusion models \cite{Sohl-Dickstein2015DeepUnsupervisedLearning, Ho2020DenoisingDiffusionProbabilistic, Bengio2013GeneralizedDenoisingAutoEncoders, Alain2016GSNsGenerativeStochastic} which gradually destroy data $\vec{x}_0$ by adding noise over a fixed number of steps $T$ using a noise schedule $\beta_{1:T}$ determined so that $\vec{x}_T$ is approximately normally distributed. The forward process is defined by a discrete Markov chain,
\begin{subequations}
\begin{align}
    q(\vec{x}_{1:T}|\vec{x}_0) &= \prod_{t=1}^T q(\vec{x}_t| \vec{x}_{t-1}), \\
    q(\vec{x}_t|\vec{x}_{t-1}) &= \mathcal{N}(\vec{x}_t ; \sqrt{1-\beta_t}\vec{x}_{t-1}, \beta_i \vec{I}).
\end{align}
\end{subequations}
The parameterised reverse process is trained to gradually remove noise, i.e. approximate $p_\theta(\vec{x}_{t-1}|\vec{x}_t)$, by optimising a re-weighted variant of the ELBO, similar to Eqn. \ref{eqn:denoising-score-matching}.

Diffusion models have also been applied to categorical data; multinomial diffusions \cite{hoogeboom2021argmax} define a forward process where each discrete variable switches randomly to a different value and the reverse process is trained to approximate the noise. Self-supervised language models such as BERT \cite{devlin2019bert} have similar training objectives: variables are randomly masked out and the model is trained to predict the original values; these models can be viewed as Markov random fields and sampled using Gibbs/Metropolis Hastings via iterative sampling of the masked distributions \cite{goyal2021exposing, wang2019bert}.

\subsubsection{Speeding up Sampling}
Sampling from score-based models requires a large number of steps leading to various techniques being developed to reduce this. A simple approach is to skip steps at inference: cosine schedules \cite{nichol2021improved} spend more time where larger visual changes are made reducing the impact of skipping; another approach is to use dynamic programming to find what steps should be taken to minimise ELBO based on a computation budget \cite{watson2021learning}.
Taking the continuous time limit of a diffusion model results in a stochastic differential equation (SDE), numerical solvers can therefore be used, reducing the number of steps required \cite{song2021scorebased, jolicoeur2021gotta}. 
Another proposed approach is to model noisy data points as $q(\vec{x}_{t-1} | \vec{x}_t, \vec{x}_0)$, allowing the generative process to skip some steps using its approximation of end samples $\vec{x}_0$ \cite{song2021denoising}.

\subsection{Correcting Implicit Generative Models}
While EBMs offer powerful representation ability due to unnormalized likelihoods, they can suffer from high variance training, long training and sampling times, and struggle to support the entire data space. In this section, a number of hybrid approaches are discussed which address these issues.

\subsubsection{Exponential Tilting}
To eliminate the need for an EBM to support the entire space, an EBM can instead be used to correct samples from an implicit generative network, simplifying the function to learn and allowing easier sampling. This procedure, referred to as exponentially tilting an implicit model, is defined as
\begin{equation}
    p_{\theta, \phi}(\vec{x}) = \frac{1}{Z_{\theta, \phi}} q_\phi(\vec{x})e^{-E_\theta(\vec{x})}.
\end{equation}
By parameterising $q_\phi(\vec{x})$ as a latent variable model such as a normalizing flow \cite{Nijkamp2020LearningEnergybasedModel, arbel2021generalized} or VAE generator \cite{xiao2021vaebm}, MCMC sampling can be performed in the latent space rather than the data space. Since the latent space is much simpler, and often uni-modal, MCMC mixes much more effectively. This limits the freedom of the model, however, leading some to jointly sample in latent and data space \cite{arbel2021generalized, xiao2021vaebm}.

\subsubsection{Noise Contrastive Estimation}
Noise contrastive estimation \cite{gutmann2010noisecontrastive, durkan2020contrastive} transforms EBM training into a classification problem using a noise distribution $q_\phi(\vec{x})$ by optimising the loss function,
\begin{equation}
    \E_{p_d} \bigg[ \ln \frac{p_\theta(\vec{x})}{p_\theta(\vec{x}) + q_\phi(\vec{x})} \bigg] + \E_{q_\phi} \bigg[ \ln \frac{q_\phi(\vec{x})}{p_\theta(\vec{x}) + q_\phi(\vec{x})} \bigg], \label{eqn:noise-contrastive-divergence}
\end{equation}
where $p_\theta(\vec{x}) = e^{E_\theta(\vec{x}) - c}$. This approach can be used to train a correction via exponential tilting \cite{Nijkamp2020LearningEnergybasedModel}, but can also be used to directly train an EBM and normalizing flow \cite{gao2020flow}. Eqn. \ref{eqn:noise-contrastive-divergence} is equivalent to GAN Equation \ref{eqn:gan-reformulation}, however, training formulations differ, with noise contrastive estimation explicitly modelling likelihood ratios.

\subsection{Alternative Training Objectives}
As aforementioned, energy models trained with contrastive divergence approximately maximises the likelihood of the data; likelihood however does not correlate directly with sample quality \cite{Theis2016NoteEvaluationGenerative}. Training EBMs with arbitrary f-divergences is possible, yielding improved FID scores \cite{yu2020training}. 

Since score estimates have high variance, the Stein discrepancy has been proposed as an alternative objective, requiring no sampling and more closely correlating with likelihood \cite{grathwohl2020learning}. A middle ground between denoising score matching and contrastive divergence is diffusion recovery likelihood \cite{bengio2013generalized} which can be optimised via a sequence of denoising EBMs conditioned on increasingly noisy samples of the data, the conditional distributions being much easier to MCMC sample from than typical EBMs \cite{gao2021learning}.

%%%%%%%%%%%%%%%%%%%%%%%%%%%%%%%%%%%%%%%%%%%%
%%%%%%%%%%%%%%%%%%%%%%%%%%%%%%%%%%%%%%%%%%%%
%%%%%%%%% Variational Autoencoders %%%%%%%%%
%%%%%%%%%%%%%%%%%%%%%%%%%%%%%%%%%%%%%%%%%%%%
%%%%%%%%%%%%%%%%%%%%%%%%%%%%%%%%%%%%%%%%%%%%

\section{Variational Autoencoders \label{sec:vae}}
One of the key problems associated with energy-based models is that sampling is not straightforward and mixing can require a significant amount of time. To circumvent this issue, it would be beneficial to explicitly sample from the data distribution with a single network pass. 

To this end, suppose we have a latent based model $p_\theta(\vec{x}|\vec{z})$ with prior $p_\theta(\vec{z})$ and posterior $p_\theta(\vec{z}|\vec{x})$; unfortunately optimising this model through maximum likelihood is intractable due to the integral in $p_\theta(\vec{x})=\int_{\vec{z}} p_\theta(\vec{x}|\vec{z})p_\theta(\vec{z})d\vec{z}$. Instead, variational inference allows this problem to be reframed as an optimisation problem by introducing an approximation of the true intractable posterior $q_\phi(\vec{z}|\vec{x}) = \arg\min_q D_{KL}(q_\phi(\vec{z}|\vec{x}) || p_\theta(\vec{z}|\vec{x}))$ that allows a tractable bound on $p_\theta(\vec{x})$ to be formed. In particular, variational autoencoders amortize the inference process, that is, approximate $q_\phi(\vec{z}|\vec{x})$ using a feedforward inference network allowing scaling to large datasets \cite{Kingma2014AutoEncodingVariationalBayes, rezende2014stochastic}. From the definition of KL divergence we get
\begin{equation}
    \begin{split}
        &D_{KL} ( q_\phi(\vec{z}|\vec{x})||p_\theta(\vec{z}|\vec{x})) = \mathbb{E}_{q_\phi(\vec{z}|\vec{x})}\bigg[ \ln \frac{q_\phi(\vec{z}|\vec{x})}{p_\theta(\vec{z}|\vec{x})} \bigg] \\
        &\quad = \mathbb{E}_{q_\phi(\vec{z}|\vec{x})}[ \ln q_\phi(\vec{z}|\vec{x}) ] - \mathbb{E}_{q_\phi(\vec{z}|\vec{x})}[ \ln p_\theta(\vec{z},\vec{x})] \\
        &\qquad + \ln p_\theta(\vec{x}),
    \end{split}
\end{equation}
which can be rearranged to find an alternative definition for $p_\theta(\vec{x})$ that does not require the knowledge of $p_\theta(\vec{z}|\vec{x})$
\begin{equation*}
    \begin{split}
        \ln p_\theta(\vec{x}) & = D_{KL}(q_\phi(\vec{z}|\vec{x})||p_\theta(\vec{z}|\vec{x})) - \mathbb{E}_{q_\phi(\vec{z}|\vec{x})}[ \ln q_\phi(\vec{z}|\vec{x}) ] \\
        &\quad + \mathbb{E}_{q_\phi(\vec{z}|\vec{x})}[ \ln p_\theta(\vec{z},\vec{x})] \\
        & \geq - \mathbb{E}_{q_\phi(\vec{z}|\vec{x})}[ \ln q_\phi(\vec{z}|\vec{x}) ] + \mathbb{E}_{q_\phi(\vec{z}|\vec{x})}[ \ln p_\theta(\vec{z},\vec{x})] \\
    \end{split}
\end{equation*}
\vspace*{-1em}
\begin{equation}
    \hspace*{3.15em}= - \mathbb{E}_{q_\phi(\vec{z}|\vec{x})}[ \ln q_\phi(\vec{z}|\vec{x}) ] + \mathbb{E}_{q_\phi(\vec{z}|\vec{x})}[ \ln p_\theta(\vec{z}) ]
\end{equation}
\vspace*{-1em}
\begin{equation*}
    \begin{split}
        \hspace*{3.5em}&\quad + \mathbb{E}_{q_\phi(\vec{z}|\vec{x})}[ \ln p_\theta(\vec{x}|\vec{z}) ] \\
        & = - D_{KL}(q_\phi(\vec{z}|\vec{x}) || p_\theta(\vec{z})) + \mathbb{E}_{q_\phi(\vec{z}|\vec{x})}[ \ln p_\theta(\vec{x}|\vec{z}) ] \\
        &\equiv \mathcal{L}(\theta, \phi; \vec{x}),
    \end{split}
\end{equation*}
where $\mathcal{L}$ is known as the evidence lower bound (ELBO) \cite{Jordan1999IntroductionVariationalMethods}. To optimise this bound with respect to $\theta$ and $\phi$, gradients must be backpropagated through the stochastic sampling process $\tilde{\vec{z}} \sim q_\phi(\vec{z}|\vec{x})$. This is permitted by reparameterizing $\tilde{\vec{z}}$ using a differentiable function $g_\phi(\vec{\epsilon}, \vec{x})$ of a noise variable $\vec{\epsilon}$: $\tilde{\vec{z}} = g_\phi(\vec{\epsilon},\vec{x})$ with $\vec{\epsilon} \sim p(\vec{\epsilon})$ \cite{Kingma2014AutoEncodingVariationalBayes, rezende2014stochastic}. 

Monte Carlo gradient estimators can be used to approximate the expectations, however, this yields very high variance making it impractical. Alternatively, if $D_{KL}(q_\phi(\vec{z}|\vec{x}) || p_\theta(\vec{z}))$ can be integrated analytically then the variance is manageable. A prior with such a property needs to be simple enough to sample from but also sufficiently flexible to match the true posterior; a common choice is a normally distributed prior with diagonal covariance, $\vec{z} \sim q_\phi(\vec{z}|\vec{x})=\mathcal{N}(\vec{z};\vec{\mu},\vec{\sigma}^2\vec{I})$ with $\tilde{\vec{z}} = \vec{\mu} + \vec{\sigma} \odot \vec{\epsilon}$ and $\vec{\epsilon} \sim \mathcal{N}(\vec{0},\vec{I})$. In this case, the loss simplifies to
\begin{equation*}
    \tilde{\mathcal{L}}_{VAE}(\theta, \phi; \vec{x}) \simeq \frac{1}{2} \sum_{j=1}^J \Big( 1 + \ln((\sigma^{(j)})^2) - (\mu^{(j)})^2 - (\sigma^{(j)})^2 \Big)
\end{equation*}
\vspace{-0.4em}
\begin{equation}
     + \frac{1}{L}\sum_{l=1}^L \ln p_\theta(\vec{x}|\tilde{\vec{z}}_l). \hspace{2.9em}
\end{equation}

Despite success on small scale datasets, when applied to more complex datasets such as natural images, samples tend to be unrealistic and blurry \cite{Dosovitskiy2016GeneratingImagesPerceptual}. This blurriness has been attributed to the maximum likelihood objective itself and MSE reconstruction loss, however, there is evidence that limited approximation of the true posterior is the root cause \cite{Zhao2017DeeperUnderstandingVariational}; with MSE causing highly non-Gaussian posteriors. As such, the Gaussian posterior implies an overly simple model which, when unable to perfectly fit, maps multiple data points to the same encoding leading to averaging. 

There are a number of other issues associated with limited posterior approximation, namely under-estimation of the variance of the posterior, resulting in poor predictions, and biases in the MAP estimates of model parameters \cite{Turner2011TwoProblemsVariational}. 
Additionally, amortized inference leads to an amortization gap, the difference in ELBO for the amortized posterior and optimal approximate posterior \cite{cremer2018inference}. Increasing the capacity of the encoder and decoder can reduce this gap by improving the posterior approximation and better fitting the choice of approximation respectively. 
Other proposed improvements include combining with adversarial training \cite{huang2018introvae, larsen2016autoencoding, makhzani2016adversarial}, improving the ELBO \cite{burda2016importance}, as well as using different regularisation such as Wasserstein distance \cite{Tolstikhin2019WassersteinAutoEncoders}.

Reweighting the ELBO by multiplying $D_{KL}$ with an extra hyperparameter $\beta$ allows the capacity of the latent representation to be altered. When $\beta>1$ a more disentangled representation is learned where each latent unit is responsible for a single generative factor \cite{higgins2017beta}. This approach has been generalised, allowing more precise states in the compression-representation trade-off to be targeted \cite{alemi2018fixing}.

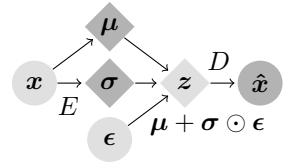
\begin{figure}[t]
    \centering
    \begin{tikzpicture}[shorten >=1pt,->]
        \tikzstyle{cir}=[circle,fill=black!30,minimum size=17pt,inner sep=0pt]
        \tikzstyle{dia}=[diamond,fill=black!30,minimum size=19pt,inner sep=0pt]
        \tikzstyle{box}=[rounded corners=6pt,fill=black!25,minimum size=17pt,inner xsep=4pt, inner ysep=0pt]
        \tikzstyle{sqr}=[fill=black!25,minimum size=17pt, inner ysep=0pt, inner xsep=4pt]
        
        \node[cir,fill=black!12]  (VAE-x)    at (0,1) {$\vec{x}$};
        \node[dia]                (VAE-mu)   at (1,1.75) {$\vec{\mu}$};
        \node[dia]                (VAE-sig)  at (1,1) {$\vec{\sigma}$};
        \node[cir,fill=black!12]  (VAE-eps)  at (1,0.25) {$\vec{\epsilon}$};
        \node[dia,fill=black!12]  (VAE-z)    at (2,1) {$\vec{z}$};
        \node[cir]                (VAE-xh)   at (3,1) {$\vec{\hat{x}}$};
        
        \draw(VAE-x) -- (VAE-mu);
        \draw(VAE-x) -- (VAE-sig);
        \draw(VAE-mu) -- (VAE-z);
        \draw(VAE-sig) -- (VAE-z);
        \draw(VAE-eps) -- (VAE-z);
        \draw(VAE-z) -- (VAE-xh);
        
        \node[align=center] at (0.45,0.7) {$E$};
        \node[align=center] at (2.45,1.3) {$D$};
        \node[align=center] at (2.3,0.45) {$\vec{\mu} + \vec{\sigma} \odot  \vec{\epsilon} $};
    \end{tikzpicture}
    \caption{Variational autoencoder with a normally distributed prior. $\vec{\epsilon}$ is sampled from $\mathcal{N}(\vec{0}, \bm{I})$.}
    \label{fig:variational-autoencoder}
\end{figure}

\subsection{Beyond Simple Priors}
One approach to improve variational bounds and increase sample quality is to improve the priors used for instance by careful selection to the task or by increasing its complexity \cite{Hoffman2016ELBOSurgeryAnother}. Complex priors can be learned by warping simple distributions and inducing variational dependencies between the latent variables: variational Gaussian processes permit this by forming an infinite ensemble of mean-field distributions \cite{tran2016variational}; EBMs and score matching can be used to model flexible priors \cite{pang2020learning, vahdat2021score}; normalizing flows (see Section \ref{sec:normalizing-flows}) transform distributions through a series of invertible parameterised functions \cite{Rezende2015VariationalInferenceNormalizing, Kingma2016ImprovedVariationalInference, Berg2018SylvesterNormalizingFlows, Grathwohl2018FFJORDFreeformContinuous,salimans2015markov,huang2017learnable}.

By rewriting the VAE training objective to have two regularisation terms \cite{makhzani2016adversarial},
\begin{equation}
\begin{split}
    \mathcal{L}(\theta, \phi; \vec{x}) &= \E_{\vec{x} \sim q(\vec{x})}[ \E_{q_\phi(\vec{z}|\vec{x})}[ \ln p_\theta(\vec{x}|\vec{z})] ] \\
    &\quad+ \E_{\vec{x} \sim q(\vec{x})}[\mathbb{H}[q_\phi(\vec{z}|\vec{x})]] - \E_{\vec{z} \sim q(\vec{z})}[p_\theta(\vec{z})],
\end{split}
\end{equation}
the latter of which is the cross entropy between the aggregate posterior and the prior, the prior can be defined as the aggregate posterior, thus obtaining a rich multi-modal latent representation that combats inactive latent variables. Since the true aggregate posterior is intractable, VampPrior \cite{tomczak2018vae} approximates it for a set of pseudo-inputs, tensors with the same shape as data points learned during training. Exemplar VAEs \cite{norouzi2020exemplar} scale this approach up, using the full training set to approximate the aggregate posterior, by approximating the prior using $k$-nearest-neighbours. Alternatively, the aggregate posterior can be approximated with a learned prior; this has been achieved with a learned rejection sampling procedure that transforms a base distribution \cite{bauer2019resampled}.

In some instances, it can be helpful to compress data to discrete latent representations \cite{bowman2016generating, kaiser2018fast}, however, gradients through discrete sampling procedures are ill-defined. The Gumbel-Softmax/Concrete distribution is a differentiable continuous approximation of a categorical distribution containing a temperature coefficient that converges to a discrete distribution in the limit \cite{jang2017categorical, maddison2017concrete}.

Alternatively, it has been argued that simple Gaussian priors are not a hindrance. When the data of dimension $d$ lies on a sub-manifold of dimension $r$ and $r<d$ then global VAE optimum exist that do no recover the data distribution, however, when $r=d$, global optimums do recover the data distribution; as such, 2 stage VAEs that first map data to latents of dimension $r$ then use a second VAE to correct the learned density can better capture the data \cite{dai2019diagnosing}.

\subsubsection{Hierarchical VAEs}
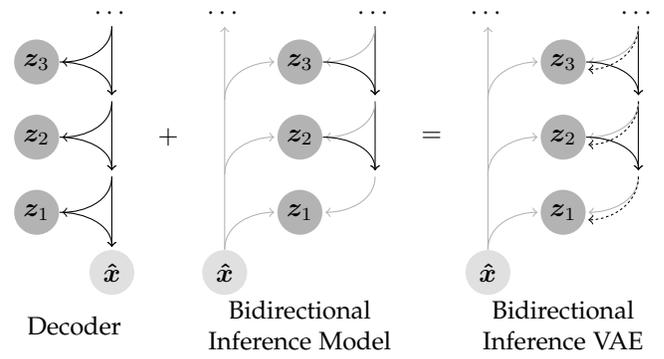
\begin{figure}[t]
\centering
\begin{tikzpicture}[shorten >=1pt,->]
    \tikzstyle{cir}=[circle,fill=black!30,minimum size=17pt,inner sep=0pt]
    \tikzstyle{dia}=[diamond,fill=black!30,minimum size=19pt,inner sep=0pt]
    \tikzstyle{box}=[rounded corners=6pt,fill=black!25,minimum size=17pt,inner xsep=4pt, inner ysep=0pt]
    \tikzstyle{sqr}=[fill=black!25,minimum size=17pt, inner ysep=0pt, inner xsep=4pt]
    
    % Generative Model
    \node[cir] (VAEIAF-G-z1) at (0,1) {$\vec{z}_1$};
    \node[cir] (VAEIAF-G-z2) at (0,2) {$\vec{z}_2$};
    \node[cir] (VAEIAF-G-z3) at (0,3) {$\vec{z}_3$};
    \node[inner sep=0pt] (VAEIAF-G-b1) at (1,3.5) {};
    \node[inner sep=0pt] (VAEIAF-G-b2) at (1,2.5) {};
    \node[inner sep=0pt] (VAEIAF-G-b3) at (1,1.5) {};
    \node[cir,fill=black!12] (VAEIAF-G-xh) at (1,0.2) {$\vec{\hat{x}}$};
    \draw (VAEIAF-G-b1) -- (VAEIAF-G-b2);
    \draw (VAEIAF-G-b2) -- (VAEIAF-G-b3);
    \draw (VAEIAF-G-b3) -- (VAEIAF-G-xh);
    \draw[->] (VAEIAF-G-b1) to [out=270,in=0] (VAEIAF-G-z3);
    \draw[->] (VAEIAF-G-z3) to [out=0,in=90] (VAEIAF-G-b2);
    \draw[->] (VAEIAF-G-b2) to [out=270,in=0] (VAEIAF-G-z2);
    \draw[->] (VAEIAF-G-z2) to [out=0,in=90] (VAEIAF-G-b3);
    \draw[->] (VAEIAF-G-b3) to [out=270,in=0] (VAEIAF-G-z1);
    \draw[->] (VAEIAF-G-z1) to [out=0,in=90] (VAEIAF-G-xh);
    \node[align=center] at (0.5,-0.5) {Decoder};
    \node[align=center] at (1,3.65) {$\cdots$};
    
    \node[align=center] at (1.75,2) {$+$};
    
    % Inference Model
    \node[cir,fill=black!12] (VAEIAF-I-xh) at (2.5,0.2) {$\vec{\hat{x}}$};
    \node[cir] (VAEIAF-I-z1) at (3.5,1) {$\vec{z}_1$};
    \node[cir] (VAEIAF-I-z2) at (3.5,2) {$\vec{z}_2$};
    \node[cir] (VAEIAF-I-z3) at (3.5,3) {$\vec{z}_3$};
    \node[inner sep=0pt] (VAEIAF-I-b1) at (4.5,3.5) {};
    \node[inner sep=0pt] (VAEIAF-I-b2) at (4.5,2.5) {};
    \node[inner sep=0pt] (VAEIAF-I-b3) at (4.5,1.5) {};
    \draw (VAEIAF-I-b1) -- (VAEIAF-I-b2);
    \draw (VAEIAF-I-b2) -- (VAEIAF-I-b3);
    \draw[->,color=black!30] (VAEIAF-I-b1) to [out=270,in=0] (VAEIAF-I-z3);
    \draw[->] (VAEIAF-I-z3) to [out=0,in=90] (VAEIAF-I-b2);
    \draw[->,color=black!30] (VAEIAF-I-b2) to [out=270,in=0] (VAEIAF-I-z2);
    \draw[->] (VAEIAF-I-z2) to [out=0,in=90] (VAEIAF-I-b3);
    \draw[->,color=black!30] (VAEIAF-I-b3) to [out=270,in=0] (VAEIAF-I-z1);
    \draw[->,color=black!30] (VAEIAF-I-xh) to (2.5,3.5);
    \draw[->,color=black!30] (VAEIAF-I-xh) to [out=90,in=180] (VAEIAF-I-z1);
    \draw[->,color=black!30] (2.5,1.5) to [out=90,in=180] (VAEIAF-I-z2);
    \draw[->,color=black!30] (2.5,2.5) to [out=90,in=180] (VAEIAF-I-z3);
    \node[align=center] at (3.5,-0.5) {Bidirectional \\ Inference Model};
    \node[align=center] at (2.5,3.65) {$\cdots$};
    \node[align=center] at (4.5,3.65) {$\cdots$};
    
    \node[align=center] at (5.25,2) {$=$};
    
    % Full VAE
    \node[cir,fill=black!12] (VAEIAF-V-xh) at (6,0.2) {$\vec{\hat{x}}$};
    \node[cir] (VAEIAF-V-z1) at (7,1) {$\vec{z}_1$};
    \node[cir] (VAEIAF-V-z2) at (7,2) {$\vec{z}_2$};
    \node[cir] (VAEIAF-V-z3) at (7,3) {$\vec{z}_3$};
    \node[inner sep=0pt] (VAEIAF-V-b1) at (8,3.5) {};
    \node[inner sep=0pt] (VAEIAF-V-b2) at (8,2.5) {};
    \node[inner sep=0pt] (VAEIAF-V-b3) at (8,1.5) {};
    \draw (VAEIAF-V-b1) -- (VAEIAF-V-b2);
    \draw (VAEIAF-V-b2) -- (VAEIAF-V-b3);
    \draw[->,color=black!30] (VAEIAF-V-b1) to [out=270,in=0] (VAEIAF-V-z3);
    \draw[->] (VAEIAF-V-z3) to [out=0,in=90] (VAEIAF-V-b2);
    \draw[->,color=black!30] (VAEIAF-V-b2) to [out=270,in=0] (VAEIAF-V-z2);
    \draw[->] (VAEIAF-V-z2) to [out=0,in=90] (VAEIAF-V-b3);
    \draw[->,color=black!30] (VAEIAF-V-b3) to [out=270,in=0] (VAEIAF-V-z1);
    \draw[->,color=black!30] (VAEIAF-V-xh) to (6,3.5);
    \draw[->,color=black!30] (VAEIAF-V-xh) to [out=90,in=180] (VAEIAF-V-z1);
    \draw[->,color=black!30] (6,1.5) to [out=90,in=180] (VAEIAF-V-z2);
    \draw[->,color=black!30] (6,2.5) to [out=90,in=180] (VAEIAF-V-z3);
    \draw[->,dashed,dash pattern=on 1pt off 1pt] (VAEIAF-V-b1) to [out=270,in=0] (7.3,2.9);
    \draw[->,dashed,dash pattern=on 1pt off 1pt] (VAEIAF-V-b2) to [out=270,in=0] (7.3,1.9);
    \draw[->,dashed,dash pattern=on 1pt off 1pt] (VAEIAF-V-b3) to [out=270,in=0] (7.3,0.9);
    \node[align=center] at (7,-0.5) {Bidirectional \\ Inference VAE};
    \node[align=center] at (6,3.65) {$\cdots$};
    \node[align=center] at (8,3.65) {$\cdots$};
    
\end{tikzpicture}
\caption{A hierarchical VAE with bidirectional inference \cite{Kingma2016ImprovedVariationalInference}.}
\label{fig:hierarchical-vae}
\end{figure}

Hierarchical VAEs build complex priors with multiple levels of latent variables, each conditionally dependent on the last, forming dependencies depthwise though the network,
\begin{subequations}
\begin{align}
    p_\theta(\vec{z}) &= p_\theta(\vec{z}_0) p_\theta(\vec{z}_1 | \vec{z}_0) \cdots p_\theta(\vec{z}_N | \vec{z}_{<N}), \\
    q_\phi(\vec{z}|\vec{x}) &= q_\phi(\vec{z}_0|\vec{x}) q_\phi(\vec{z}_1|\vec{z}_0, \vec{x}) \cdots q_\phi(\vec{z}_N|\vec{z}_{<N}, \vec{x}).
\end{align}
\end{subequations}
Ladder VAEs \cite{sonderby2016ladder} achieve this conditioning structure using a bidirectional inference network where a deterministic ``bottom-up'' pass generates features at various resolutions, then the latent variables are processed from top to bottom with the features shared (Fig. \ref{fig:hierarchical-vae}). Specifically, they model latents as normal distributions conditioned on the last latent,
\begin{align}
    p_\theta(\vec{z}_i|\vec{z}_{i-1}) &= \mathcal{N}(\vec{z}_i | \mu_{p,i}(\vec{z}_{i-1}), \sigma^2_{p,i}(\vec{z}_{i-1})). 
\end{align}
By introducing skip connections around the stochastic sampling process, latents can be conditioned on all previously sampled latents \cite{Kingma2016ImprovedVariationalInference, maaloe2019biva, vahdat2020nvae}. Such an architecture generalises autoregressive models; inferring latents in parallel allows for significantly fewer steps compared to typical autoregressive models since many latents are statistically independent and allows different latent levels to correspond to global/local details depending on their depth. It has been argued that a single level of latents is sufficient since Gibbs sampling performed on that level can recover the data distribution \cite{zhao2017learning}. Despite that, Gibbs sampling converges slowly, making hierarchical representations more efficient; in support of this, deeper hierarchical VAEs have been shown to improve likelihood, independent of capacity \cite{child2021very}.

\subsection{Regularised Autoencoders}
Related to VAEs are regularised autoencoders (RAEs) which apply regularisation to the latent space of a deterministic autoencoder then subsequently train a density estimator on this space to obtain a complex prior \cite{ghosh2020variational}. Since the approximate posterior is a degenerate distribution, RAEs have little connection with variational inference. Vector Quantized-Variational Autoencoders (VQ-VAE) \cite{VanDenOord2017NeuralDiscreteRepresentation, ramesh2021zeroshot} achieve this by training an autoencoder with a discrete latent space, then approximating encodings with an autoregressive model (see Section \ref{sec:autoregressive}). The encoder's outputs are compared to a codebook of latent vectors and set to the code they are closest to; the gradient of this discretisation process is approximated using the straight through estimator \cite{Bengio2013EstimatingPropagatingGradients}. Meanwhile, latent vectors in the codebook are moved closer to the encoder's outputs. To model larger images, hierarchy of codes have been applied \cite{Razavi2019GeneratingDiverseHighFidelity}, as well as adversarial learning to increase compression rate \cite{esser2021taming}.

\subsection{Data Modelling Distributions}
Unlike energy-based models, VAEs must model an explicit density $p(\vec{x}|\vec{z})$. For efficient sampling, typically this distribution is decomposed as a product of independent simple distributions, allowing unrestricted architectures to be used to parameterise the chosen distributions. Common instances include modelling variables as Bernoulli \cite{loaiza-ganem2019continuous}, Gaussian \cite{Kingma2014AutoEncodingVariationalBayes}, multinomial distributions, or as mixtures\cite{Salimans2017PixelCNNImprovingPixelCNN}.

\subsubsection{Autoregressive Decoders}
To introduce dependencies between the output variables, numerous works have used powerful autoregressive networks \cite{gulrajani2016pixelvae}. While these approaches allow complex distributions to be learned, they increase the runtime and often suffer from posterior collapse since early in training the approximate posterior contains little knowledge about $\vec{x}$ meaning that it is easy to minimise $D_{KL}$ which in turn reduces the gradient between the encoder and decoder making it difficult to escape this minima \cite{bowman2016generating}; in fact, for a sufficiently powerful generative distribution, this can occur even at optimum solutions \cite{Chen2017VariationalLossyAutoencoder}.
Various methods to prevent posterior collapse have been proposed: by restricting the autoregressive network's receptive field to a small window, it is forced to use latents to capture global structure \cite{Chen2017VariationalLossyAutoencoder}; a mutual information term can be added to the loss to encourage high correlation between $\vec{x}$ and $\vec{z}$ \cite{zhao2019infovae}; encouraging the posterior to be diverse by controlling its geometry to evenly covering the data space, redundancy is reduced and latents are encouraged to learn global structure \cite{ma2019mae}.

\subsection{Bridging Amortized and Stochastic Inference} 
While variational approaches offer substantial speedup over MCMC sampling, there is an inherent discrepancy between the true posterior and approximate posterior despite improvements in this field. To this end, a number of approaches have been proposed to find a middle ground, yielding improvements over amortized methods with lower costs than MCMC. Semi-amortised VAEs \cite{kim2018semiamortized} use an encoder network followed by stochastic gradient descent on latents to improve the ELBO, however, this still relies on an inference network. The inference network can be removed by assigning latent vectors to data points, then optimising them with Langevin dynamics or gradient descent, during training; although this allows fast training, convergence for unseen samples is not guaranteed and there is still a large discrepancy between the true posterior and latent approximations due to lag in optimisation \cite{han2017alternating, bojanowski2019optimizing}. Short-run MCMC has also been applied however it has poor mixing properties \cite{nijkamp2020learninga}. Gradient Origin Networks \cite{Bond-Taylor2020GradientOriginNetworks} replace the encoder with an empirical Bayes approximation of the posterior that only requires a single gradient step.

VAEBMs offer a different perspective, rather than performing latent MCMC sampling based on the ELBO, they use an auxiliary energy-based model to correct blurry VAE samples, with MCMC sampling performed in both the data space and latent space. This setup is defined by $h_{\phi,\theta}(\vec{x}, \vec{z}) = \frac{1}{Z_{\phi,\theta}}p_\theta(\vec{z})p_\theta(\vec{x}|\vec{z})e^{-E_\phi(\vec{x})}$, where $p_\theta(\vec{z})p_\theta(\vec{x}|\vec{z})$ is the VAE, and $E_\phi(\vec{x})$ is the energy model. This, however, requires 2 stages of training to avoid calculating the gradient of the normalising constant $Z_{\phi,\theta}$, training only the VAE and fixing the VAE and training the EBM respectively.

%%%%%%%%%%%%%%%%%%%%%%%%%%%%%%%%%%%%%%%%%%%%
%%%%%%%%%%%%%%%%%%%%%%%%%%%%%%%%%%%%%%%%%%%%
%%%%%%% GENERATIVE ADVERSARIAL NETS %%%%%%%%
%%%%%%%%%%%%%%%%%%%%%%%%%%%%%%%%%%%%%%%%%%%%
%%%%%%%%%%%%%%%%%%%%%%%%%%%%%%%%%%%%%%%%%%%%

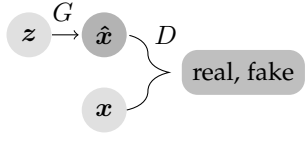
\begin{figure}[t]
    \centering
    \begin{tikzpicture}[shorten >=1pt,->]
        % define a couple of points for drawing
        \tikzstyle{cir}=[circle,fill=black!30,minimum size=17pt,inner sep=0pt]
        \tikzstyle{dia}=[diamond,fill=black!30,minimum size=19pt,inner sep=0pt]
        \tikzstyle{box}=[rounded corners=6pt,fill=black!25,minimum size=17pt,inner xsep=4pt, inner ysep=0pt]
        \tikzstyle{sqr}=[fill=black!25,minimum size=17pt, inner ysep=0pt, inner xsep=4pt]
          
        \node[cir,fill=black!12]  (GAN-z)  at (0,1) {$\vec{z}$};
        \node[cir]  (GAN-xh) at (1,1) {$\vec{\hat{x}}$};
        \node[cir,fill=black!12]  (GAN-x)  at (1,0) {$\vec{x}$};
        \node[box]  (GAN-rf) at (2.85,0.5) {real, fake};
        \draw(GAN-z) -- (GAN-xh);
        \draw[>=, decorate, decoration={brace, amplitude=16pt, mirror}] (1.3,0) -- coordinate [left=10pt] (B) (1.3,1) node {};
        
        \node[align=center] at (0.45,1.3) {$G$};
        \node[align=center] at (1.82,1.0) {$D$};
    \end{tikzpicture}
    \caption{Generative adversarial networks set two networks in a game: $D$ detects real from fake samples while $G$ tricks $D$.}
    \label{fig:generative-adversarial-network}
\end{figure}

\section{Generative Adversarial Networks \label{sec:gan}}
Another approach at eliminating the Markov chains used in energy models is the generative adversarial network (GAN) \cite{Goodfellow2014GenerativeAdversarialNets}. GANs consist of two networks, a discriminator $D \colon \reals^n \to [0,1]$ which estimates the probability that a sample comes from the data distribution $\vec{x} \sim p_d(\vec{x})$, and a generator $G \colon \reals^m \to \reals^n$ which given a latent variable $\vec{z} \sim p_{\vec{z}}(\vec{z})$, captures $p_d$ by tricking the discriminator into thinking its samples are real. This is achieved through adversarial training of the networks: $D$ is trained to correctly label training samples as real and samples from $G$ as fake, while $G$ is trained to minimise the probability that $D$ classifies its samples as fake. This can be interpreted as $D$ and $G$ playing a mini-max game, as with prior work \cite{schmidhuber1990making, schmidhuber2020generative}, optimising the value function $V(G,D)$,
\begin{equation}
\begin{split}
    \min_G \max_D V(G,D) &= \E_{\vec{x}\sim p_d(\vec{x})}[\ln D(\vec{x})] \\
    &\quad + \E_{\vec{z}\sim p_{\vec{z}}(\vec{z})}[\ln(1-D(G(\vec{z})))].
\end{split}
\end{equation}
For a fixed $G$, the objective for $D$ can be reformulated as
\begin{equation*}
    \max_D V(G,D) = \E_{\vec{x}\sim p_d}[\ln D(\vec{x})] + \E_{\vec{x}\sim p_{g}}[\ln(1-D(\vec{x}))]
\end{equation*}
\begin{equation}
\begin{split}
    &= \E_{\vec{x}\sim p_d}\bigg[ \ln \frac{p_d(\vec{x})}{p_d(\vec{x}) + p_g(\vec{x})} \bigg] \hspace*{9.3em} \\
    &\quad + \E_{\vec{x}\sim p_g}\bigg[ \ln \frac{p_{g}(\vec{x})}{p_d(\vec{x}) + p_g(\vec{x})} \bigg]
    \label{eqn:gan-reformulation}
\end{split}
\end{equation}
\begin{equation*}
    = D_{KL}(p_d || \tfrac{1}{2}(p_d + p_g)) + D_{KL}(p_g ||  \tfrac{1}{2}(p_d + p_g)) + C.
\end{equation*}
Therefore the loss is equivalent to the Jensen-Shannon divergence between the generative distribution $p_g$ and the data distribution $p_d$ and thus with sufficient capacity, the generator can recover the data distribution. The use of symmetric JS-divergence is well behaved when both distributions are small unlike the asymmetric KL-divergence used in maximum likelihood models. Additionally, it has been suggested that reverse KL-divergence, $D_{KL}(p_g || p_d)$, is a better measure for training generative models than normal KL-divergence, $D_{KL}(p_d || p_g)$, since it minimises $\E_{\vec{x} \sim p_g}[\ln p_d(\vec{x})]$ \cite{Huszar2015HowNotTrain}; while reverse KL-divergence is not a viable objective function, JS-divergence is and behaves more like reverse KL-divergence than KL-divergence alone. With that said, JS-divergence is not perfect; if 0 mass is associated with a data sample in a maximum likelihood model, KL-divergence is driven to infinity, whereas this can happen with no consequence in a GAN.

\begin{figure*}[t]
    \centering
    \begin{subfigure}{0.78\textwidth}
    \vspace{0.4em}
      \begin{tabular}{>{\raggedright}p{21mm}p{66mm}p{42mm}}
        \toprule
        Name         & Discriminator Loss         & Generator Loss \\
        \midrule
        NSGAN \cite{Goodfellow2014GenerativeAdversarialNets}   & $-\E[\ln(\sigma(D(\vec{x})))] - \E[\ln(1-\sigma(D(G(\vec{z}))))]$ & $-\E[\ln(\sigma(D(G(\vec{z}))))]$ \\
        
        WGAN \cite{Arjovsky2017WassersteinGAN} & $\E[D(\vec{x})] - \E[D(G(\vec{z}))]$ & $\E[D(G(\vec{z}))]$ \\
        
        LSGAN \cite{Mao2017LeastSquaresGenerative} & $\E[(D(\vec{x})-1)^2] + \E[D(G(\vec{z}))^2]$ & $\E[(D(G(\vec{z}))-1)^2]$ \\
        
        Hinge \cite{lim2017geometric} & $\E[\min(0,D(\vec{x})-1)] - \E[\max(0,1+D(G(\vec{z})))]$ & $-\E[D(G(\vec{z}))]$ \\
        
        EBGAN \cite{Zhao2017EnergybasedGenerativeAdversarial} & $D(\vec{x})+\max(0,m-D(G(\vec{z})))$ & $D(G(\vec{z}))$ \\
        
        RSGAN \cite{Jolicoeur-Martineau2018RelativisticDiscriminatorKey} & $\E[\ln(\sigma(D(\vec{x})-D(G(\vec{z}))))]$ & $\E[\ln(\sigma(-D(G(\vec{z}))-D(\vec{x})))]$ \\
        \bottomrule
      \end{tabular}
      \vspace{0.4em}
      \caption{GAN losses.}
      \label{tab:gan-losses}
    \end{subfigure}%
    \begin{subfigure}{0.22\textwidth}
        \centering
        \begin{adjustbox}{width=\textwidth}
            \begin{tikzpicture}[font=\footnotesize, outer sep=0.1em, inner sep=0]
            \begin{axis}[
                axis lines = left,
                xlabel = $D(G(\vec{z}))$,
                ylabel = {$\loss_G$},
                ytick distance=2,
                legend cell align={left},
                ymin=-2.5, ymax=9.5,
                width=\textwidth,
                ytick=\empty,
                yticklabels={,,},
                xtick=\empty,
                xticklabels={,,},
                %label style={font=\smallfont},
                %legend style={nodes=right},
                scale only axis
            ]
            \addplot+[
                very thick,
                domain=-2:2, 
                samples=100, 
                color=myblue,
                mark options={fill=myblue},
                mark repeat=100
            ]
            {ln(1-(1/(1+(e^(-x)))))};
            \addlegendentry{S-GAN}
            
            \addplot+[
                very thick,
                domain=-2:2, 
                samples=100, 
                color=myorange,
                mark options={fill=myorange},
                mark repeat=100
            ]
            {-ln(1/(1+(e^(-x))))};
            \addlegendentry{NS-GAN}
            
            \addplot+[
                very thick,
                domain=-2:2, 
                samples=100, 
                color=mygreen,
                mark options={fill=mygreen},
                mark repeat=100,
                mark=triangle
            ]
            {x};
            \addlegendentry{WGAN}
            
            \addplot+[
                very thick,
                domain=-2:2, 
                samples=100, 
                color=myred,
                mark options={fill=myred},
                mark repeat=100
            ]
            {(x-1)^2};
            \addlegendentry{LS-GAN}
            
            \addplot+[
                very thick,
                domain=-2:2, 
                samples=100, 
                color=mypurple,
                mark options={fill=mypurple},
                mark repeat=100
            ]
            {-x};
            \addlegendentry{Hinge}
        
            \end{axis}
            \end{tikzpicture}
        \end{adjustbox}
        \caption{Generator loss functions.}
        \label{fig:gan-losses}
    \end{subfigure}
    \caption{A comparison of popular losses used to train GANs. (a) Respective losses for discriminator/generator. (b) Plots of generator losses with respect to discriminator output. Notably, NS-GAN's gradient disappears as discriminator gets better.}
\end{figure*}
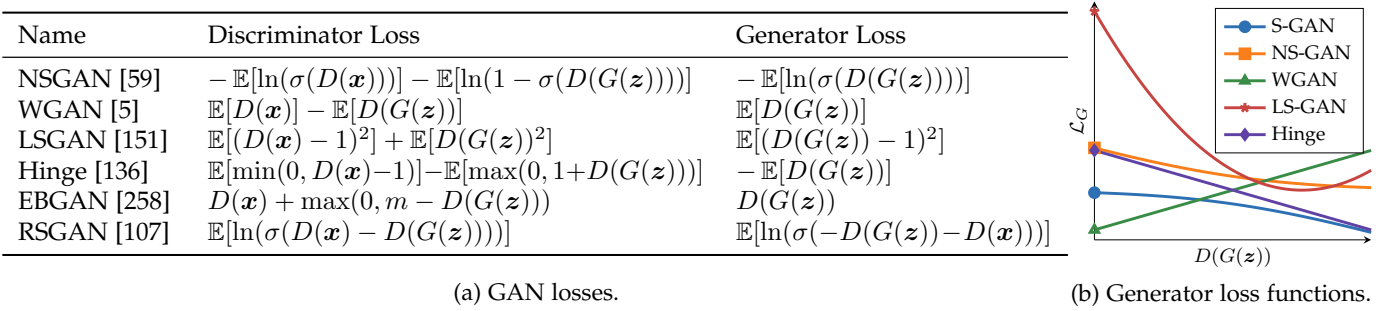

\subsection{Stabilising Training}
The adversarial nature of GANs makes them notoriously difficult to train \cite{arjovsky2017principled}; Nash equilibrium is hard to achieve \cite{Salimans2016ImprovedTechniquesTraining} since non-cooperation cannot guarantee convergence, thus training often results in oscillations of increasing amplitude. As the discriminator improves, gradients passed to the generator vanish, accelerating this problem; on the other hand, if the discriminator remains poor, the generator does not receive useful gradients. Another problem is mode collapse, where one network gets stuck in a bad local minima and only a small subset of the data distribution is learned. The discriminator can also jump between modes resulting in catastrophic forgetting, where previously learned knowledge is forgotten when learning something new \cite{thanh-tung2020catastrophic}. This section explores proposed solutions to these problems.

\subsubsection{Loss Functions}
Since the cause of many of these issues can be linked with the use of JS-divergence, other loss functions have been proposed that minimise other statistical distances; in general, any $f$-divergence can be used to train GANs \cite{Nowozin2016FGANTrainingGenerative}. One notable example is the Wasserstein distance which intuitively indicates how much ``mass'' must be moved to transform one distribution into another. Wasserstein distance is defined formally in Eqn. \ref{eqn:wasserstein-distance}, which by the Kantorovich-Rubinstein duality is equivalent to Eqn. \ref{eqn:wasserstein-duality} \cite{villani2008optimal}:
\begin{subequations}
\begin{align}
    W(p_d, p_g) &= \inf_{\gamma \in \prod(p_d,p_g)} \E_{(\vec{x},\vec{y})\sim\gamma}[\hspace{2px} \norm{\vec{x}-\vec{y}} ] \label{eqn:wasserstein-distance}, \\
    W(p_d,p_g) &= \sup_{\norm{D}_L \leq 1} \E_{\vec{x} \sim p_d}[D(\vec{x})] - \E_{\vec{x} \sim p_g}[D(\vec{x})], \label{eqn:wasserstein-duality}
\end{align}
\end{subequations}
where the supremum is taken over all 1-Lipschitz functions, that is, $f$ such that for all $x_1$ and $x_2$, $\norm{f(x_1)-f(x_2)}_2 \leq  \norm{x_1 - x_2}_2$. Optimising Wasserstein distance, as described in Table \ref{tab:gan-losses}, offers linear gradients thus eliminating the vanishing gradients problem (see Fig. \ref{fig:gan-losses}). Moreover, Wasserstein distance is also equivalent to minimising reverse KL-divergence \cite{Miyato2018SpectralNormalizationGenerative}, offers improved stability, and  allows training to optimality. Numerous approaches to enforce 1-Lipschitz continuity have been proposed: weight clipping \cite{Arjovsky2017WassersteinGAN} invalidates gradients making optimisation difficult; applying a gradient penalty within the loss is heavily dependent on the support of the generative distribution and computation with finite samples makes application to the entire space intractable \cite{Gulrajani2017ImprovedTrainingWasserstein}; spectral normalisation (discussed below) applies global regularisation by estimating the singular values of parameters. Other popular loss functions include least squares GAN, hinge loss, energy-based GAN, and relativistic GAN (detailed in Table \ref{tab:gan-losses}).

The catastrophic forgetting problem can be mitigated by conditioning the GAN on class information, encouraging more stable representations \cite{mirza2014conditional, Zhang2019SelfAttentionGenerativeAdversarial, Brock2019LargeScaleGAN}. Nevertheless, labelled data, if available, only covers limited abstractions. Self-supervision achieves the same goal by training the discriminator on an auxiliary classification task based solely on the unsupervised data. Proposed approaches are based on randomly rotating inputs to the discriminator, which learns to identify the angle rotated separately to the standard real/fake classification \cite{chen2019selfsupervised}. Extensions include training the discriminator to jointly determine rotation and real/fake to provide better feedback \cite{tran2019selfsupervised}, and training the generator to trick the discriminator at both the real/fake and classification tasks \cite{tran2019selfsupervised}. A more explicit approach is to model the generator with a normalizing flow, avoiding collapse by jointly optimising the GAN and likelihood objectives \cite{Grover2018FlowGANCombiningMaximum}.

\subsubsection{Spectral Normalisation}
Spectral normalisation \cite{Miyato2018SpectralNormalizationGenerative} is a technique to make a function globally 1-Lipschitz utilising the observation that the Lipschitz constant of a linear function is its largest singular value (spectral norm). The spectral norm of a matrix $\vec{A}$ is 
\begin{equation}
    SN(\vec{A}) := \max_{\vec{h}:\vec{h}\neq\vec{0}} \frac{\norm{\vec{A}\vec{h}}_2}{\norm{\vec{h}}_2} = \max_{\norm{\vec{h}}_2 \leq 1} \norm{\vec{A}\vec{h}}_2,
\end{equation}
thus a weight matrix $\vec{W}$ is normalised to be 1-Lipschitz by replacing the weights with $\vec{W}_{SN} := \frac{\vec{W}}{SN(\vec{W})}$. Rather than using singular value decomposition to compute the norm, the power iteration method is used; for randomly initialised vectors $\vec{v} \in \reals^n$ and $\vec{u} \in \reals^m$, the procedure is
\begin{equation}
    \vec{u}_{t+1} = \vec{W}\vec{v}_t, \hspace{0.35em}  \vec{v}_{t+1} = \vec{W}^T\vec{u}_{t+1}, \hspace{0.35em} SN(\vec{W}) \approx \vec{u}^T\vec{W}\vec{v}.
\end{equation}
Since weights change only marginally with each optimisation step, a single power iteration step per global optimisation step is sufficient to keep $\vec{v}$ and $\vec{u}$ close to their targets.% values. 

As aforementioned, enforcing the discriminator to be 1-Lipschitz is essential for WGANs, however, spectral normalisation has been found to dramatically improve sample quality and allow scaling to datasets with thousands of classes across a variety of loss functions \cite{Miyato2018SpectralNormalizationGenerative, Brock2019LargeScaleGAN}. Spectral collapse, has been linked to discriminator overfitting when spectral norms of layers explode \cite{Brock2019LargeScaleGAN} as well as mode collapse when spectral norms fall in value significantly \cite{Liu2019SpectralRegularizationCombating}. Additionally, regularising the discriminator in this manner helps balance the two networks, reducing the number of discriminator update steps required \cite{Zhang2019SelfAttentionGenerativeAdversarial, Brock2019LargeScaleGAN}.

\subsubsection{Data Augmentation}\label{sec:gan-augmentation}
Augmenting training data to increase the quantity of training data is often common practice; when training GANs the types of augmentations permitted are limited to more simple augmentations such as cropping and flipping to prevent the generator from creating undesired artefacts. Several approaches independently proposed applying augmentations to all discriminator inputs, allowing more substantial augmentations to be used \cite{zhao2020differentiable, zhao2020image, tran2021data, karras2020training}; the training procedure for a WGAN with augmentations is
\begin{subequations}
\begin{align}
    \mathcal{L}_D &= \E_{\vec{x} \sim p_d(\vec{x})}[D(T(\vec{x}))] - \E_{\vec{z} \sim p(\vec{z})}[D(T(G(\vec{z})))], \\
    \mathcal{L}_G &= \E_{\vec{z} \sim p(\vec{z})}[D(T(G(\vec{z})))],
\end{align}
\end{subequations}
where $T$ is a random augmentation. These approaches have been shown to improve sample quality on equivalent architectures and stabilise training. Each work offers a different perspective on why augmentation is so effective: the increased quantity of training data in conjunction with the more difficult discrimination task prevents overfitting and in turn collapse \cite{Brock2019LargeScaleGAN}, notably this applies even on very small datasets (100 samples); the nature of GAN training leads to the generated and data distributions having non-overlapping supports, complicating training \cite{sonderby2017amortised}, strong augmentations may cause these distributions to overlap further. If an augmentation is differentiable and represents an invertible transformation of the data space's distribution, then the JS-divergence is invariant, and the generator is guaranteed to not create augmented samples \cite{tran2021data, karras2020training}.

\subsubsection{Discriminator Driven Sampling}
In order to improve sample quality and address overpowered discriminators, numerous works have taken inspiration from the connection between GANs and energy models \cite{Zhao2017EnergybasedGenerativeAdversarial}. Interpreting the discriminator of a Wasserstein GAN \cite{Arjovsky2017WassersteinGAN} as an energy-based model means samples from the generator can be used to initialise an MCMC sampling chain which converges to the density learned by the discriminator, correcting errors learned by the generator \cite{turner2019metropolishastings, neklyudov2019implicit}. This is similar to pure EBM approaches, however, training the two networks adversarially changes the dynamics. The slow convergence rates of high dimensional MCMC sampling has led others to instead sample in the latent space \cite{che2020your, song2020discriminator}.

\subsubsection{GANs without Competition}
Originally proposed as a proxy to measure GAN convergence \cite{grnarova2019domain}, the duality gap is an upper bound on the JS-divergence that can be directly optimised \cite{grnarova2021generative}, defined as
\begin{equation}
    DG(D,G) = \max_{D'} V(G,D') - \min_{G'} V(G',D).
\end{equation}
Cooperative training simplifies the optimisation procedure, avoiding oscillations. Each training step, however, requires optimising for $D'$ and $G'$ which slows down training and could suffer from vanishing gradients.

\subsection{Architectures}
Careful network design is a key component for stable GAN training. Scaling any deep neural network to high-resolution data is non-trivial due to vanishing gradients and high memory usage, but since the discriminator can classify high-resolution data more easily, GANs notably struggle \cite{odena2017conditional}.

Early approaches designed hierarchical architectures, dividing the learning procedure into more easily learnable chunks. LapGAN \cite{denton2015deep} builds a Laplacian pyramid such that at each layer, a GAN conditioned on the previous image resolution predicts a residual adding detail.  Stacked GANs \cite{huang2017stacked, Zhang2017StackGANTextPhotoRealistic} use two GANs trained successively: the first generates low-resolution samples, then the second upsamples and corrects the first, thus fewer GANs need to be trained. A related approach, progressive growing \cite{Karras2018ProgressiveGrowingGANsa, Karras2019StyleBasedGeneratorArchitecture}, iteratively trains a single GAN at higher resolutions by adding layers to both the generator and discriminator upscaling the previous output, after the previous resolution converges. Training in this manner, however, not only takes a long time but leads to high frequency components being learned in the lower layers, resulting in shift artefacts \cite{Karras2020AnalyzingImprovingImage}.

Accordingly, a number of works have targeted a single GAN that can be trained end-to-end. DCGAN \cite{Radford2016UnsupervisedRepresentationLearning} introduced a fully convolutional architecture with batch normalisation \cite{Ioffe2015BatchNormalizationAccelerating} and ReLU/LeakyReLU activations. BigGAN \cite{Brock2019LargeScaleGAN} employ a number of tricks to scale to high resolutions including using very large mini-batches to reduce variation, spectral normalisation to discourage spectral collapse, and using large datasets to prevent overfitting. Despite this, training collapse still occurs thus requiring early stopping. Another approach is to include skip connections between the generator and discriminator at each resolution, allowing gradients to flow through shorter paths to each layer, providing extra information to the generator \cite{karnewar2020msggan, Karras2020AnalyzingImprovingImage, valvano2021learning}. By treating subsets of the generator's parameters as smaller generators, Anycost GANs extend this approach, allowing samples to be generated at multiple resolutions and speeds\cite{lin2021anycost}. To learn long-range dependencies, GANs can be built with self-attention components \cite{Vaswani2017AttentionAllYou, Zhang2019SelfAttentionGenerativeAdversarial, jiang2021transgan}, however, full quadratic attention does not scale well to high dimensional data.

\subsection{Training Speed}
The mini-max nature of GAN training leads to slow convergence, if achieved at all. This problem has been exacerbated by numerous works as a byproduct of improving stability or sample quality. One such example is that by using very large mini-batches, reducing variance and covering more modes, sample quality can be improved significantly, however, this comes at the cost of slower training  \cite{Brock2019LargeScaleGAN}. Small-GAN \cite{sinha2020smallgan} combats this by replacing large batches with small batches that approximate the shape of the larger batch using core set sampling \cite{sinha2020smallgan}, significantly improving the mode coverage and sample quality of GANs trained with small batches.

While strong discriminator regularisation stabilises training, it allows the generator to make small changes and trick the discriminator, making convergence very slow. Rob-GAN \cite{liu2019robgan}, include an adversarial attack step \cite{madry2019deep} that perturbs real images to trick the discriminator without altering the content inordinately, adapting the GAN objective into a min-max-min problem. This provides a weaker regularisation, enforcing small Lipschitz values locally rather than globally. This approach has been connected with the follow-the-ridge algorithm \cite{zhong2020improving, wang2020solving}, an optimisation approach for solving mini-max problems that reduces the optimisation path and converges to local mini-max points.

Another approach to improve training speed is to design more efficient architectures. Depthwise convolutions \cite{chollet2017xception} apply separate convolutions to each channel of a tensor reducing the number of operations and hence also the runtime, have been found to have comparable quality to standard convolutions \cite{ngxande2019depthwisegans}. Lightweight GANs \cite{liu2021faster} achieve fast training using a number of tricks including small batch sizes, skip-layer excitation modules which provide efficient shortcut gradient flow, as well as using a self-supervised discriminator forcing good features to be learned.

%%%%%%%%%%%%%%%%%%%%%%%%%%%%%%%%%%%%%%%%%%%%
%%%%%%%%%%%%%%%%%%%%%%%%%%%%%%%%%%%%%%%%%%%%
%%%%%%%%%% AUTOREGRESSIVE MODELS %%%%%%%%%%%
%%%%%%%%%%%%%%%%%%%%%%%%%%%%%%%%%%%%%%%%%%%%
%%%%%%%%%%%%%%%%%%%%%%%%%%%%%%%%%%%%%%%%%%%%

\section{Autoregressive Likelihood Models \label{sec:autoregressive}}

Autoregressive generative models \cite{Bengio2003NeuralProbabilisticLanguage} are based on the chain rule of probability, where the probability of a variable that can be decomposed as $\vec{x} = x_1,\hdots,x_n$ is expressed as
\begin{equation}
    p(\vec{x}) = p(x_1,\hdots,x_n) = \prod_{i=1}^n p(x_i|x_1,\hdots,x_{i-1}).
\end{equation}
As such, unlike GANs and energy models, it is possible to directly maximise the likelihood of the data by training a recurrent neural network to model $p(x_i|\vec{x}_{1:i-1})$ by minimising the negative log-likelihood,
\begin{equation}
    -\ln p(\vec{x}) = - \sum_i^n \ln p(x_i|x_1,\hdots,x_{i-1}).
\end{equation}
While autoregressive models are extremely powerful density estimators, sampling is inherently a sequential process and can be exceedingly slow on high dimensional data. Additionally, data must be decomposed into a fixed ordering; while the choice of ordering can be clear for some modalities (e.g. text and audio), it is not obvious for others such as images and can affect performance depending on the network architecture used.

\subsection{Architectures}
The majority of research is focused on improving network architectures to increase their receptive fields and memory, ensuring the network has access to all parts of the input to encourage consistency, as well as increasing the network capacity, allowing more complex distributions to be modelled. 

\subsubsection{Masked Multilayer Perceptrons}
One approach to build autoregressive models is to mask the weights of simple multilayer perceptron (MLP) autoencoders so as to satisfy the autoregressive property. The neural autoregressive density estimator (NADE) \cite{larochelle2011neural}, which can be viewed as a mean-field approximation of a restricted Boltzmann machine, achieves this for binary data by placing time-dependent masks on an MLP with one hidden layer. Specifically, at time step $i$, weights are masked so that the entire hidden state $\vec{h}_i$ and output $p(x_i | \vec{x}_{<i})$ are dependent only on $\vec{x}_{<i}$; formally this can be defined as
\begin{subequations}
\begin{align}
    p(x_i=1 | \vec{x}_{<i}) &= \sigma(b_i + (\vec{W}^T)_{i,\cdot} \vec{h}_i), \\
    \vec{h}_i &= \sigma(\vec{c} + \vec{W}_{\cdot,<i} \vec{x}_{<i}),
\end{align}
\end{subequations}
where $\vec{W}_{\cdot,<d}$ is the first $d-1$ columns of a shared weight matrix $\vec{W}$, and $b_i$ and $\vec{c}$ are biases. The RNADE \cite{Uria2013RNADE} generalises NADE to real valued data by instead modelling $p(x_i | \vec{x}_{<i})$ with mixture distributions parameterised by the network. An alternative masking procedure known as MADE \cite{germain2015made} allows for parallel density estimation by placing a mask fixed over time on an MLP so that no connections exist between $p(x_i| \vec{x}_{<i})$ and $\vec{x}_{\geq i}$. Additionally, MADE is more readily vectorisable and does not suffer from neuron saturation since the number of inputs to all neurons is constant with respect to time.

\subsubsection{Recurrent Neural Networks}
A natural architecture to apply is that of standard recurrent neural networks (RNNs) such as LSTMs \cite{Hochreiter1997LongShortTermMemory, theis2015generative, VanDenOord2016PixelRecurrentNeural} and GRUs \cite{Chung2014EmpiricalEvaluationGated, Mehri2017SampleRNNUnconditionalEndtoEnd} which model sequential data by tracking information in a hidden state. However, RNNs are known to forget information, limiting their receptive field thus preventing modelling of long range relationships. This can be improved by stacking RNNs that run at different frequencies allowing long data such as multiple seconds of audio to be modelled \cite{Chung2014EmpiricalEvaluationGated}. Nevertheless, their sequential nature means that training can be too slow for many tasks. 

\subsubsection{Causal Convolutions}
An alternative approach is that of causal convolutions, which apply masked or shifted convolutions over a sequence \cite{VanDenOord2016ConditionalImageGeneration,Salimans2017PixelCNNImprovingPixelCNN, Chen2017PixelSNAILImprovedAutoregressive}. When stacked, this only provides a receptive field linear with depth, however, by dilating the convolutions to skip values with some step the receptive field can be orders of magnitude higher. 

\subsubsection{Self-Attention}
Neural attention is an approach which at each successive time step is able to select where it wishes to `look' at previous time steps. This concept has been used to autoregressively `draw' images onto a blank `canvas' \cite{Gregor2015DRAWRecurrentNeural} in a manner similar to human drawing. More recently self-attention (known as Transformers when used in an encoder-decoder setup) \cite{Vaswani2017AttentionAllYou} has made significant strides improving not only autoregressive models, but also other generative models due to its parallel nature, stable training, and ability to effectively learn long-distance dependencies. This is achieved using an attention scheme that can reference any previous input where an entirely independent process is used per time step so that there are no dependencies. Specifically, inputs are encoded as key-value pairs, where the values $\bm{V}$ represent the inputs, and the keys $\bm{K}$ act as an indexing method. At each time step a query $q$ is made; taking the dot product of the queries and keys, a similarity vector is formed that describes which value vectors to access. This process can be expressed as
\begin{equation}
    \text{Attention}(\bm{Q},\bm{K},\bm{V}) = \text{softmax}\bigg( \frac{\bm{Q}\bm{K}^T}{\sqrt{d_k}} \bigg)\bm{V},
\end{equation}
where $d_k$ is the key/query dimension and is used to normalise gradient magnitudes. Since the self-attention process contains no recurrence, positional information must be passed into the function. A simple effective method to achieve this is to add sinusoidal positional encodings which combine sine and cosine functions of different frequencies to encode positional information \cite{Vaswani2017AttentionAllYou}; alternatively others use trainable positional embeddings \cite{Child2019GeneratingLongSequences}.

The infinite receptive fields of attention provides a powerful tool for representing data, however, the attention matrix $\bm{Q}\bm{K}^T$ grows quadratically with data dimension, making scaling difficult. Approaches include scaling across large quantities of GPUs \cite{Brown2020LanguageModelsAre}, interleaving attention between causal convolutions \cite{Chen2017PixelSNAILImprovedAutoregressive}, attending over local regions \cite{parmar2018image}, and using sparse attention patterns that provide global attention when multiple layers are stacked \cite{Child2019GeneratingLongSequences}. More recently, a number of linear transformers have been proposed whose memory and time footprints grow linearly with data dimension \cite{katharopoulos2020transformers, choromanski2021rethinking, wang2020linformer}. By approximating the softmax operation with a kernel function with feature representation $\phi(\vec{x})$, the order of multiplications can be rearranged to
\begin{equation}
    \big( \phi(\bm{Q}) \phi(\bm{K})^T \big) \bm{V} = \phi(\bm{Q}) \big( \phi(\bm{K})^T \bm{V} \big),
\end{equation}
allowing $\phi(\bm{K})^T \bm{V}$ to be cached and used for each query.

\begin{figure}[t]
    \centering
    \begin{tikzpicture}[shorten >=1pt,->]
        \tikzstyle{cir}=[circle,fill=black!30,minimum size=17pt,inner sep=0pt]
        \tikzstyle{dia}=[diamond,fill=black!30,minimum size=19pt,inner sep=0pt]
        \tikzstyle{box}=[rounded corners=6pt,fill=black!25,minimum size=17pt,inner xsep=4pt, inner ysep=0pt]
        \tikzstyle{sqr}=[fill=black!25,minimum size=17pt, inner ysep=0pt, inner xsep=4pt]
        
        \node[cir,fill=black!12]  (A-1)    at (0,0) {$x_1$};
        \node[cir,fill=black!12]  (A-2)    at (1,0) {$x_2$};
        \node[cir,fill=black!12]  (A-3)    at (2,0) {$x_3$};
        \node[align=center] at (2.75,0) {$...$};
        \node[rounded corners=16pt,box,fill=black!12] (A-4) at (3.75,0) {$x_{n-1}$};
        
        \node[cir]  (A-5)  at (0,1.3) {$\hat{x}_1$};
        \node[cir]  (A-6)  at (1,1.3) {$\hat{x}_2$};
        \node[cir]  (A-7)  at (2,1.3) {$\hat{x}_3$};
        \node[align=center]  at (2.75,1.3) {$...$};
        \node[rounded corners=16pt,box,inner xsep=9pt] (A-8) at (3.75,1.3) {$\hat{x}_n$};
        
        \foreach \from/\to in {1/5,1/6,1/7,1/8,2/6,2/7,2/8,3/7,3/8,4/8}
        { \draw (A-\from) -- (A-\to); }
    \end{tikzpicture}
    \caption{Autoregressive models decompose data points using the chain rule and learn conditional probabilities.}
    \label{fig:autoregressive-model}
\end{figure}
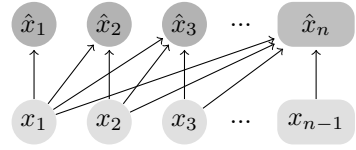

\subsubsection{Multiscale Architectures}
Even with a linear autoregressive model, $O(N)$ for $N$ pixels, scaling to high-resolution images grows quadratically with resolution. One multi-scale approach reduces this complexity to $O(\ln N)$ by successively upscaling images, making the assumption that when upscaling, each pixel is dependent only on its adjacent area and the previous resolution image, allowing scaling to high resolutions \cite{reed2017parallel}. To avoid making independence assumptions, \cite{menick2019generating} partition images in an interleaving pattern so that sub-images are the same size and capture global structure. Sub-images are generated autoregressively pixel-wise and are conditioned on previously generated sub-images; while this reduces the memory required, sampling times are still slow.

\subsection{Data Modelling Decisions}
When generating text, output variables are often modelled using a multinomial distribution since tokens are discrete and are in general unrelated. However, this modelling assumption can cause complications or be infeasible in other cases such as 16-bit audio modelling, in which magnitude would not be intrinsically modelled and 65,536 output neurons would be required. Solutions proposed include:
\begin{itemize}
    \item Applying $\mu$-law, a logarithmic companding algorithm which takes advantage of human perception of sound, then quantizing to 8-bit values \cite{Oord2016WaveNetGenerativeModel}.
    \item First predicting the first 8-bits, then predicting the second 8-bits conditioned on the first. 
    \item Modelling output probabilities using a mixture of logistic distributions (MoL) has the benefits of providing more useful gradients and allowing intensities never seen to still be sampled \cite{Salimans2017PixelCNNImprovingPixelCNN}.
\end{itemize}
Nevertheless, these assumptions restrict the expressiveness of the network, for instance, MoLs struggle to model high frequency signals as found in raw image data; a simple solution in this case is to add Gaussian noise, reducing the Lipschitz constant of the data distribution \cite{meng2021improved}. This restriction can be removed at the expense of less efficient sampling by learning an autoregressive energy model, for instance, by approximating normalising constants \cite{nash2019autoregressive} or through score matching \cite{meng2020autoregressive}. Alternatively, quantile regression, which minimises Wasserstein distance, can be used to learn an approximation of the inverse cumulative distribution \cite{Ostrovski2018AutoregressiveQuantileNetworks}.

When modelling images, many works use ``raster scan'' ordering \cite{VanDenOord2016PixelRecurrentNeural, VanDenOord2016ConditionalImageGeneration, Salimans2017PixelCNNImprovingPixelCNN} where pixels are estimated row by row. Alternatives have been proposed such as ``zig-zag'' ordering \cite{Chen2017PixelSNAILImprovedAutoregressive} which allows pixels to depend on previously sampled pixels to the left and above, providing more relevant context. Another factor when modelling images is how to factorise sub-pixels. While it is possible to treat them as independent variables, this adds additional complexity. Alternatively, it is possible to instead condition on whole pixels, and output joint distributions in a single step \cite{Salimans2017PixelCNNImprovingPixelCNN}.

%%%%%%%%%%%%%%%%%%%%%%%%%%%%%%%%%%%%%%%%%%%%
%%%%%%%%%%%%%%%%%%%%%%%%%%%%%%%%%%%%%%%%%%%%
%%%%%%%%%%%% NORMALIZING FLOWS %%%%%%%%%%%%%
%%%%%%%%%%%%%%%%%%%%%%%%%%%%%%%%%%%%%%%%%%%%
%%%%%%%%%%%%%%%%%%%%%%%%%%%%%%%%%%%%%%%%%%%%

\begin{table*}[t]
  \caption{Normalizing Flow Layers: $\odot$ represents elementwise multiplication, $\star_l$ represents a cross-correlation layer}
  \label{tab:normalizing-flows}
  \centering
  \begin{tabular}{>{\raggedright}p{30mm}p{46mm}p{49mm}p{40mm}<{\raggedright}}%p{30mm}<{\raggedright}}
    \toprule
    Description         & Function          & Inverse Function & Log-Determinant \\ %& Properties \\
    
    \midrule
    Low Rank & \\
    \midrule
    
    Planar \cite{Rezende2015VariationalInferenceNormalizing}\newline & 
    $\vec{y} = \vec{x} + \vec{u}h(\vec{w}^T\vec{z}+b)$ \newline With $\vec{w} \in \mathbb{R}^D$, $\vec{u} \in \mathbb{R}^D$, $\vec{b} \in \mathbb{R}$ \newline & 
    No closed form inverse &
    $\ln | 1 + \vec{u}^T h'(\vec{w}^T\vec{z} + b)\vec{w} |$\newline \\ %& Difficult to invert \newline Single unit bottleneck \\
    
    Sylvester \cite{Hasenclever2017VariationalInferenceOrthogonal, Berg2018SylvesterNormalizingFlows} & 
    $\vec{y} = \vec{x} + \vec{U}h(\vec{W}^T\vec{x}+\vec{b})$ & 
    No closed form inverse &
    $\ln\det(\vec{I}_M + \text{diag}(h'(\vec{W}^T\vec{x}+\vec{b}))\vec{W}\vec{U}^T)$ \\
    
    \midrule
    Coupling/Autoregressive &  \\
    \midrule
    
    General Coupling \newline\newline & 
    $\vec{y}^{(1:d)} = \vec{x}^{(1:d)}$ \newline $\vec{y}^{(d+1:D)} = h(\vec{x}^{(d+1:D)}; f_\theta(\vec{x}^{(1:d)}))$ & 
    $\vec{x}^{(1:d)} = \vec{y}^{(1:d)}$ \newline $\vec{x}^{(d+1:D)} = h^{-1}(\vec{y}^{(d+1:D)}; f_\theta(\vec{y}^{(1:d)}))$ &
    $\ln | \det \nabla_{\vec{x}^{(d+1:D)}} h |$ \\
    
    % TODO: Check log-determinants and inverses for MAF and IAF
    MAF \cite{Papamakarios2017MaskedAutoregressiveFlow}\newline & 
    $y^{(t)} = h(x^{(t)}; f_\theta(\vec{x}^{(1:t-1)}))$ &
    $x^{(t)} = h^{-1}(y^{(t)}; f_\theta(\vec{x}^{(1:t-1)}))$ & 
    $-\sum_{t=1}^D \ln |\frac{\partial y^{(t)}}{\partial x^{(t)}}|$\\
    
    IAF \cite{Kingma2016ImprovedVariationalInference}\newline & 
    $y^t = h(x^{(t)}; f_\theta(\vec{y}^{(1:t-1)}))$ &
    $x^t = h^{-1}(y^{(t)}; f_\theta(\vec{y}^{(1:t-1)}))$ & 
    $\sum_{t=1}^D \ln |\frac{\partial y^{(t)}}{\partial x^{(t)}}|$\\
    
    Affine Coupling \cite{Dinh2016DensityEstimationUsing}\newline & 
    $h(\vec{x}; \vec{\theta}) = \vec{x} \odot \exp(\vec{\theta}_1) + \vec{\theta}_2$ & 
    $h^{-1}(\vec{y};\vec{\theta}) = (\vec{y} - \vec{\theta}_2) \odot \exp(-\vec{\theta}_1) $ &
    $\sum_{i=1}^d \theta_1^{(i)}$ \\
    
    Flow++ \cite{ho2019flow}\newline\newline & 
    $h(\vec{x}; \vec{\theta}) = \exp(\vec{\theta}_1) \odot F(\vec{x}, \vec{\theta}_3) + \vec{\theta}_2$ \newline where $F$ is a monotone function. &
    Calculated through bisection search &
    $\sum_{i=1}^d \theta_1^{(i)} + \ln \frac{\partial F(\vec{x}, \vec{\theta}_3)_i}{\partial x_i}$ \\
    
    Spline Flows \cite{Muller2019NeuralImportanceSampling}\newline\cite{durkan2019cubicspline}\cite{Durkan2019NeuralSplineFlows}\newline & 
    $h(\vec{x}; \vec{\theta}) = \text{Spline}(\vec{x}; \vec{\theta})$ \newline where $\vec{\theta}$ are the spline's knots. &
    $h^{-1}(\vec{y}; \vec{\theta}) = \text{Spline}^{-1}(\vec{y}; \vec{\theta})$ &
    Computed in closed-form as a product of quotient derivatives \\
    
    B-NAF \cite{DeCao2019BlockNeuralAutoregressive} \newline & 
    $\vec{y}=\vec{W}\vec{x}^T$ for blocked weights: \newline
    $\vec{W} = \exp(\tilde{\vec{W}}) \odot \vec{M}_d + \tilde{\vec{W}} \odot \vec{M}_o$ \newline
    where $\vec{M}_d$ selects diagonal blocks and $\vec{M}_o$ selects off-diagonal blocks. & 
    No closed form inverse & % Check this is correct
    $\ln \sum_{i=1}^d \exp(\tilde{W}_{ii})$ \\
    
    \midrule
    Convolutions & \\
    \midrule
    
    1x1 Convolution \cite{Kingma2018GlowGenerativeFlow} \newline\newline & 
    $h \times w \times c$ tensor $\vec{x}$ \& $c \times c$ tensor $\vec{W}$ \newline $\forall i,j : \vec{y}_{i,j} = \vec{W}\vec{x}_{i,j}$ & $\forall i,j : \vec{x}_{i,j} = \vec{W}^{-1}\vec{y}_{i,j}$
    & $h \cdot w \cdot \ln|\det \vec{W}|$ \\
    
    Emerging \newline Convolutions \cite{Hoogeboom2019EmergingConvolutionsGenerative} & 
    $\vec{k} = \vec{w}_1 \odot \vec{m}_1$, \quad $\vec{g} = \vec{w}_2 \odot \vec{m}_2$ \newline $\vec{y} = \vec{k} \star_l (\vec{g} \star_l \vec{x})$ & 
    $\vec{z}_t = (\vec{y}_t - \sum_{i=t+1} G_{t,i} \vec{z}_i) / G_{t,t}$ \newline $\vec{x}_t = (\vec{z}_t - \sum_{i=1}^{t-1} K_{t,i} \vec{x}_i) / K_{t,t}$ & % check correctness
    $\sum_c \ln | \vec{k}_{c,c,m_y,m_x} \vec{g}_{c,c,m_y,m_x} |$ \\
    
    \midrule
    Lipshitz Residual & \\
    \midrule
    
    i-ResNet \cite{Behrmann2019InvertibleResidualNetworks}\newline & 
    $\vec{y} = \vec{x} + f(\vec{x})$ \newline where $\norm{f}_L < 1$ & 
    $\vec{x}_1 = \vec{y}. \quad$ $\vec{x}_{n+1} = \vec{y} - f(\vec{x}_n)$ \newline converging at an exponential rate &
    $\text{tr}(\ln(\vec{I} + \nabla_{\vec{x}} f)) = \sum_{k=1}^\infty (-1)^{k+1} \frac{\text{tr}((\nabla_{\vec{x}} f)^k)}{k}$ \\ %& Sum converges in 5-10 steps. Uses Hutchinsons trace estimator. \\
    
    \bottomrule
  \end{tabular}
\end{table*}

\begin{figure}[t]
\centering
\begin{tikzpicture}[shorten >=1pt,->]
    \tikzstyle{dcir}=[circle,draw=black,minimum size=42pt,dashed,line width=0.75pt,dash pattern=on 2pt off 2pt]
    \tikzstyle{arr}=[fill=black!12, single arrow, draw=none]
    
    \tikzstyle{cir}=[circle,fill=black!30,minimum size=17pt,inner sep=0pt]
    \tikzstyle{dia}=[diamond,fill=black!30,minimum size=19pt,inner sep=0pt]
    \tikzstyle{box}=[rounded corners=6pt,fill=black!25,minimum size=17pt,inner xsep=4pt, inner ysep=0pt]
    \tikzstyle{sqr}=[fill=black!25,minimum size=17pt, inner ysep=0pt, inner xsep=4pt]
    
    % Big dashed circles
    \node[dcir] (NF-z0dash) at (0,0) {};
    \node[dcir] (NF-zidash) at (4,0) {};
    \node[dcir] (NF-zkdash) at (6.6,0) {};
    
    % Arrows + dots
    % \node[arr, minimum width=50pt] {arrow};
    % \node[arr] (NF-z0) to (NF-z1);
    \node[inner sep=0pt] at (0.8,0.8) {\includegraphics[width=35pt]{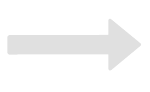}};
    \node[inner sep=0pt] at (2.6,0.8) {\includegraphics[width=35pt]{imgs/flow/arrow.pdf}};
    \node[inner sep=0pt] at (4.8,0.8) {\includegraphics[width=35pt]{imgs/flow/arrow.pdf}};
    \node[inner sep=0pt] at (3.47,0.8) {\includegraphics[width=21pt]{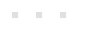}};
    \node[inner sep=0pt] at (5.85,0.8) {\includegraphics[width=21pt]{imgs/flow/dots.pdf}};
    
    % Grey z circles
    \node[cir] (NF-z0) at (0,0.8) {$\vec{x}_0$};
    \node[cir] (NF-z1) at (1.8,0.8) {$\vec{x}_1$};
    \node[cir] (NF-zi) at (4,0.8) {$\vec{x}_k$};
    \node[cir] (NF-zk) at (6.6,0.8) {$\vec{x}_K$};
    
    % Distributions
    \node[align=center] at (0,-1) {$\vec{x}_0 \sim p_0(\vec{x}_0)$};
    \node[align=center] at (4,-1) {$\vec{x}_k \sim p_k(\vec{x}_k)$};
    \node[align=center] at (6.6,-1) {$\vec{x}_K \sim p_K(\vec{x}_K)$};
    
    % Flow functions
    \node[align=center] at (0.9,1.3) {$f_1(\vec{x}_0)$};
    \node[align=center] at (2.7,1.3) {$f_2(\vec{x}_1)$};
    \node[align=center] at (5,1.3) {$f_{k+1}(\vec{x}_k)$};
    
    % Graphs
    \node[inner sep=0pt] at (0,0) {\includegraphics[width=30pt]{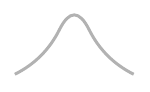}};
    \node[inner sep=0pt] at (4,0) {\includegraphics[width=30pt]{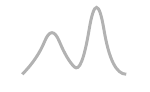}};
    \node[inner sep=0pt] at (6.6,0) {\includegraphics[width=30pt]{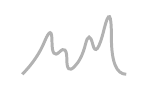}};
    
    % Graph axes
    \draw[black, line width=0.5pt] (-0.5,-0.2) -- (0.5,-0.2);
    \draw[black, line width=0.5pt] (0,-0.4) -- (0,0.45);
    
    \draw[black, line width=0.5pt] (3.5,-0.2) -- (4.5,-0.2);
    \draw[black, line width=0.5pt] (4,-0.4) -- (4,0.45);
    
    \draw[black, line width=0.5pt] (6.1,-0.2) -- (7.1,-0.2);
    \draw[black, line width=0.5pt] (6.6,-0.4) -- (6.6,0.45);
    
    \node[align=center] at (7.3,0.8) {$=\vec{y}$};
\end{tikzpicture}
\caption{Normalizing flows build complex distributions by mapping a simple distribution through invertible functions.}
\label{fig:normalizing-flow}
\end{figure}

\section{Normalizing Flows \label{sec:normalizing-flows}}
While training autoregressive models through maximum likelihood offers plenty of benefits including stable training, density estimation, and a useful validation metric, the slow sampling speed and poor scaling properties handicaps them significantly. Normalizing flows are a technique that also allows exact likelihood calculation while being efficiently parallelisable as well as offering a useful latent space for downstream tasks.
Consider an invertible, smooth function $f \colon \reals^d \to \reals^d$; by applying this transformation to a random variable $\vec{x} \sim p(\vec{x})$, then the distribution of the resulting random variable $\vec{y}=f(\vec{x})$ can be determined through the change of variables rule (and application of the chain rule),
\begin{equation} \label{eqn:flow-change-variables}
    p(\vec{y}) = p(\vec{x}) \bigg| \det \frac{\partial f^{-1}}{\partial \vec{y}} \bigg| = p(\vec{x}) \bigg| \det \frac{\partial f}{\partial \vec{x}} \bigg|^{-1}.
\end{equation}
Consequently, arbitrarily complex densities can be constructed by composing simple maps and applying Eqn. \ref{eqn:flow-change-variables} \cite{villani2003topics}. This chain is known as a normalizing flow \cite{Rezende2015VariationalInferenceNormalizing} (see Fig. \ref{fig:normalizing-flow}). The density $p_K(\vec{x}_K)$ obtained by successively transforming a random variable $\vec{x}_0$ with distribution $p_0$ through a chain of $K$ transformations $f_k$ can be defined as
\begin{subequations}
\begin{align}
    \vec{x}_K & = f_K \circ \cdots \circ f_2 \circ f_1(\vec{x}_0), \\
    \ln p_K(\vec{x}_K) & = \ln p_0(\vec{x}_0) - \sum_{k=1}^K \ln \bigg| \det \frac{\partial f_k}{\partial \vec{x}_{k-1}} \bigg|.
\end{align}
\end{subequations}
Each transformation therefore must be sufficiently expressive while being easily invertible and have an efficient to compute Jacobian determinant. While restrictive, there have been a number of works which have introduced more powerful invertible functions (see Table \ref{tab:normalizing-flows}). Nevertheless, normalizing flow models are typically less parameter efficient than other generative models.

\begin{figure}[t]
    \centering
    \begin{tikzpicture}[shorten >=1pt]
        \tikzstyle{cir}=[circle,fill=black!30,minimum size=17pt,inner sep=0pt]
        \tikzstyle{dia}=[diamond,fill=black!30,minimum size=19pt,inner sep=0pt]
        \tikzstyle{box}=[rounded corners=6pt,fill=black!25,minimum size=17pt,inner xsep=4pt, inner ysep=0pt]
        \tikzstyle{sqr}=[fill=black!25,minimum size=17pt, inner ysep=0pt, inner xsep=4pt]
        
        \node[cir,fill=black!12]  (R-11)    at (0,0) {$x_1$};
        \node[cir,fill=black!12]  (R-12)    at (1,0) {$x_2$};
        \node[cir,fill=black!12]  (R-13)    at (2,0) {$x_3$};
        \node[cir,fill=black!12]  (R-14)    at (3,0) {$x_4$};
        
        \node[dia]  (R-21)    at (0,1) {$z_1$};
        \node[dia]  (R-22)    at (1,1) {$z_2$};
        \node[cir,fill=black!12]  (R-23) at (2,1) {}; \node[] at (2,1) {$\scriptstyle h_3^{(1)}$};
        \node[cir,fill=black!12]  (R-24) at (3,1) {}; \node[] at (3,1) {$\scriptstyle h_4^{(1)}$};
        
        \node[dia]  (R-33)    at (2,2) {$z_3$};
        \node[cir,fill=black!12]  (R-34) at (3,2) {}; \node[] at (3,2) {$\scriptstyle h_4^{(2)}$};
        
        \node[dia]  (R-41)  at (0,3) {$z_1$};
        \node[dia]  (R-42)  at (1,3) {$z_2$};
        \node[dia]  (R-43)  at (2,3) {$z_3$};
        \node[dia]  (R-44)  at (3,3) {$z_4$};
        
        \foreach \from/\to in {11/21,11/22,11/23,11/24,12/21,12/22,12/23,12/24,13/21,13/22,13/23,13/24,14/21,14/22,14/23,14/24}
        { \draw[->] (R-\from) -- (R-\to); }
        
        \draw[->] (R-23) -- (R-33); \draw[->] (R-24) -- (R-34);
        \draw[->] (R-23) -- (R-34); \draw[->] (R-24) -- (R-33);
        \draw[->] (R-34) -- (R-44);
        \draw[->,densely dashed] (R-21) -- (R-41);
        \draw[->,densely dashed] (R-22) -- (R-42);
        \draw[->,densely dashed] (R-33) -- (R-43);
    \end{tikzpicture}
    \caption{Factoring out variables at different scales allows normalizing flows to scale to high dimensional data.}
    \label{fig:multiscale-flow}
\end{figure}
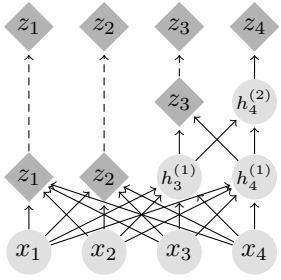

One disadvantage of requiring transformations to be invertible is that the input dimension must be equal to the output dimension which makes deep models inefficient and difficult to train. A popular solution to this is to use a multi-scale architecture \cite{Dinh2016DensityEstimationUsing, Kingma2018GlowGenerativeFlow} (see Fig. \ref{fig:multiscale-flow}) which divides the process into a number of stages, at the end of each half of the remaining units are factored out and treated immediately as outputs. This allows latent variables to sequentially represent course to fine features and permits deeper architectures.

\subsection{Coupling and Autoregressive Layers}
A simple way of building an expressive invertible function is the coupling flow \cite{Dinh2015NICENonlinearIndependent}, which divide inputs into two and applies a bijection $h$ on one half parameterised by the other,
\vspace*{-1em}
\begin{subequations}
    \begin{align}
        \vec{y}^{(1:d)} &= \vec{x}^{(1:d)}, \\
        \vec{y}^{(d+1:D)} &= h(\vec{x}^{(d+1:D)}; f_\theta(\vec{x}^{(1:d)})),
    \end{align}
\end{subequations}
here $f$ can be arbitrarily complex i.e. a neural network. $h$ tends to be selected as an elementwise function making the Jacobian triangular allowing efficient computation of the determinant, i.e. the product of elements on the diagonal.

\subsubsection{Affine Coupling}
A simple example of this is the affine coupling layer \cite{Dinh2016DensityEstimationUsing},% defined as 
\begin{equation}
     \vec{y}^{(d+1:D)} = \vec{x}^{(d+1:D)} \odot \exp(f_\sigma(\vec{x}^{(1:d)})) + f_\mu(\vec{x}^{(1:d)}),
\end{equation}
which has a simple Jacobian determinant and can be trivially rearranged to obtain a definition of $\vec{x}^{(d+1:D)}$ in terms of $\vec{y}$, provided that the scaling coefficients are not 0. This simplicity, however, comes at the cost of expressivity; while stacking numerous such flows increases their expressivity, allowing them to learn representations of complex high dimensional data such as images \cite{Kingma2018GlowGenerativeFlow}, it is unknown whether multiple affine flows are universal approximators \cite{Papamakarios2019NormalizingFlowsProbabilistic}.

\subsubsection{Monotone Functions}
Another method of creating invertible functions that can be applied element-wise is to enforce monotonicity. One possibility to achieve this is to define $h$ as an integral over a positive but otherwise unconstrained function $g$ \cite{wehenkel2019unconstrained},
\begin{equation}
    h(x_i; \vec{\theta}) = \int_0^{x_i} g_\phi(x;\vec{\theta}_1) dx + \theta_2,
\end{equation}
however, this integration requires numerical approximation. Alternatively, by choosing $g$ to be a function with a known integral solution, $h$ can be efficiently evaluated. This has been accomplished using positive polynomials \cite{Jaini2019SumofSquaresPolynomialFlow} and the CDF of a mixture of logits \cite{ho2019flow}. Both cases, however, don't have analytical inverses and have to be approximated iteratively with bisection search. Another option is to represent $g$ as a monotonic spline: a piecewise function where each piece is easy to invert. As such, the inverse is as fast to evaluate as the forward pass. Linear and quadratic splines \cite{Muller2019NeuralImportanceSampling}, cubic splines \cite{durkan2019cubicspline}, and rational-quadratic splines \cite{Durkan2019NeuralSplineFlows} have been applied so far.

\subsubsection{Autoregressive Flows}
For a single coupling layer, a significant proportional of inputs remain unchanged. A more flexible generalisation of coupling layers is the autoregressive flow, or MAF \cite{Papamakarios2017MaskedAutoregressiveFlow},
\begin{equation}
    y^{(t)} = h(x^{(t)}; f_\theta(\vec{x}^{(1:t-1)})).
\end{equation}
Here $f_\theta$ can be arbitrarily complex, allowing the use of advances in autoregressive modelling (Section. \ref{sec:autoregressive}), and $h$ is a bijection as used for coupling layers. Some monotonic bijectors have been created specifically for autoregressive flows, namely Neural Autoregressive Flows (NAF) \cite{Huang2018NeuralAutoregressiveFlows} and Block NAF \cite{DeCao2019BlockNeuralAutoregressive}. Unlike coupling layers, a single autoregressive flow is a universal approximator.

Alternatively, an autoregressive flow can be conditioned on $\vec{y}^{(1:t-1)}$ rather than $\vec{x}^{(1:t-1)}$, this is known as an Inverse Autoregressive Flow, or IAF \cite{Kingma2016ImprovedVariationalInference}. While coupling layers can be evaluated efficiently in both directions, MAF permits parallel density estimation but sequential sampling, and IAF permits parallel sampling but sequential density estimation.

\subsubsection{Probability Density Distillation}
Inverse autoregressive flows \cite{Kingma2016ImprovedVariationalInference} offer the ability to sample from an autoregressive model in parallel, however, training via maximum likelihood is inherently sequential making this infeasible for high dimensional data. Probability density distillation \cite{Oord2018ParallelWaveNetFast} has been proposed as a solution to this where a second pre-trained autoregressive network is used as a `teacher' network while an IAF network is used as a `student' and mimics the teacher's distribution by minimising the KL divergence between the two distributions:
\begin{equation}
    D_{KL}(p_S || p_T) = H(p_S, p_T) - H(p_S),
\end{equation}
where $p_S$ and $p_T$ are the student's and teacher's distributions respectively, $H(p_S, p_T)$ is the cross-entropy between $p_S$ and $p_T$, and $H(p_S)$ is the entropy of $p_S$. Crucially, this never requires the student's inverse function to be used allowing it to be computed entirely in parallel.

\subsection{Convolutional}
A considerable problem with coupling and autoregressive flows is the restricted triangular Jacobian, meaning that all inputs cannot interact with each other. Simple solutions involve fixed permutations on the output space such as reversing the order \cite{Dinh2015NICENonlinearIndependent,Dinh2016DensityEstimationUsing}. A more general approach is to use a $1\times1$ convolution which is equivalent to a linear transformation applied across channels \cite{Kingma2018GlowGenerativeFlow}. Numerous works have been proposed to generalise these to larger kernel sizes. A number of these apply variations on causal convolutions \cite{Oord2016WaveNetGenerativeModel}, including emerging convolutions \cite{Hoogeboom2019EmergingConvolutionsGenerative} whose inverse is sequential, MaCow \cite{ma2019macow} which uses smaller conditional fields allowing more efficient sampling, and MintNet \cite{song2019mintnet} which approximates the inverse using fixed-point iteration. Alternative approaches to causal masking involve imposing repeated (periodic) structure  \cite{karami2019invertible}, however in general this is not a good assumption for image modelling, as well as representing convolutions as exponential matrix-vector products, $\exp(\vec{M})\vec{x}$, approximated implicitly with a power series, allowing otherwise unconstrained kernels \cite{hoogeboom2020convolution}.

\subsection{Residual Flows}
Residual networks \cite{he2016deep} are a popular technique to build deep neural networks that alleviate the vanishing gradients problem. By restricting $f_\theta$, invertible residual networks can be built by stacking blocks of the form
\begin{equation}
    \vec{y} = \vec{x} + f_\theta(\vec{x}).
\end{equation}

\subsubsection{Matrix Determinant Lemma}
If a function has a certain residual form, then its Jacobian determinant can be computed with the matrix determinant lemma \cite{Rezende2015VariationalInferenceNormalizing}. A simple example is planar flow \cite{Rezende2015VariationalInferenceNormalizing} which is equivalent to a 3 layer MLP with a single neuron bottleneck:
\begin{equation}
    \vec{y} = \vec{x} + \vec{u} h(\vec{w}^T \vec{x} + b),
\end{equation}
where $\vec{u}, \vec{w} \in \reals^d$, $b \in \reals$, and $h$ is a differentiable non-linearity function. Planar flows are invertible provided some simple conditions are satisfied, however its inverse is difficult to compute making it only practical for density estimation tasks. A higher rank generalisation of the matrix determinant lemma has been applied to planar flows, known as Sylvester flows, removing the severe bottleneck thus allowing greater representation ability \cite{Hasenclever2017VariationalInferenceOrthogonal, Berg2018SylvesterNormalizingFlows}.

\subsubsection{Lipschitz Constrained}
By restricting the Lipschitz constant of $f_\theta$, $\norm{f_\theta}_L < 1$, then this block is invertible \cite{Behrmann2019InvertibleResidualNetworks}. The inverse, however, has no closed form definition but can be found through fixed-point iteration which by the Banach fixed-point theorem converges to a fixed unique solution at an exponential rate dependant on $\norm{f_\theta}_L$. The authors originally proposed a biased approximation of the log determinant of the Jacobian as a power series where the Jacobian trace is approximated using Hutchkinson's trace estimator (see Table \ref{tab:normalizing-flows}), but an unbiased approximator known as a Russian roulette estimator has also been proposed \cite{chen2019residual}. Unlike coupling layers, residual flows have dense Jacobians, allowing interaction. Enforcing Lipschitz constraints has been achieved with convolutional networks \cite{Miyato2018SpectralNormalizationGenerative, Liu2019SpectralRegularizationCombating, gouk2021regularisation} as well as self-attention \cite{kim2020lipschitz}.

Making strong Lipschitz assumptions severely restricts the class of functions learned; an $N$ layer residual flow network is at most $2^N$-Lipshitz. Implicit flows \cite{lu2021implicit} bypass this by solving implicit equations of the form
\begin{equation}
    F(\vec{x}, \vec{y}) = f_\theta(\vec{x}) - f_\phi(\vec{y}) + \vec{x} - \vec{y} = \vec{0},
\end{equation}
where both $f_\theta$ and $f_\phi$ both have Lipschitz constants less than 1. Both the forwards (solve for $\vec{y}$ given $\vec{x}$) and backwards (solve for $\vec{x}$ given $\vec{y}$) directions require solving a root finding problem similar to the inverse process of residual flows; indeed, an implicit flow is equivalent to the composition of a residual flow and the inverse of a residual flow. This allows them to model arbitrary Lipschitz transformations.

\subsection{Surjective and Stochastic Layers}
Restricting the class of functions available to those that are invertible introduces a number of practical problems related to the topology-preserving property of diffeomorphisms. For example, mapping a uni-modal distribution to a multi-modal distribution is extremely challenging, requiring a highly varying Jacobian \cite{Dinh2019RADApproachDeep}. By composing bijections with surjective or stochastic layers these topological constraints can be bypassed \cite{nielsen2020survae}. While the log-likelihood of stochastic layers can only be bounded by their ELBO, functions surjective in the inference direction permit exact likelihood evaluation even with altered dimensionality. Surjective transformations have the following likelihood contributions:
\begin{equation}
    \mathbb{E}_{q(\vec{y}|\vec{x})} \bigg[ \ln \frac{p(\vec{x}|\vec{y})}{q(\vec{y}|\vec{x})} \bigg],
\end{equation}
where $p(\vec{x}|\vec{y})$ is deterministic for generative surjections, and $q(\vec{y}|\vec{x})$ is deterministic for inference surjections.

One approach to build a surjective layer is to augment the input space with additional dimensions allowing smoother transformation to be learned \cite{Dupont2019AugmentedNeuralODEs, huang2020augmented, chen2020vflow}; the inverse process, where some dimensions are factored out, is equivalent to a multi-scale architecture \cite{Dinh2016DensityEstimationUsing}. Another approach known as RAD \cite{Dinh2019RADApproachDeep} learns a partitioning of the data space into disjoint subsets $\{\mathcal{Y}_i\}_{i=1}^K$, and applies piece-wise bijections to each region $g_i \colon \mathcal{X} \to \mathcal{Y}_i, \forall i \in \{1,\dots,K\}$. The generative direction learns a classifier on $\mathcal{X}$, $i \sim p(i|\vec{x})$, allowing the inverse to be calculated as $\vec{y} = g_i(\vec{x})$. Similar to both of these approaches are CIFs \cite{cornish2020relaxing} which consider a continuous partitioning of the data space via augmentation equivalent to an infinite mixture of normalizing flows. Other approaches include modelling finite mixtures of flows \cite{duan2020transport}.

Some powerful stochastic layers have already been discussed in this survey, namely VAEs \cite{Kingma2014AutoEncodingVariationalBayes} and DDPMs \cite{Ho2020DenoisingDiffusionProbabilistic}. Stochastic layers have been incorporated into normalizing flows by interleaving small energy models, sampled with MCMC, between bijectors \cite{wu2020stochastic}.

\subsection{Discrete Flows}
The normalizing flow framework can be extended to discrete distributions, by restricting transformation functions to be discrete e.g. $f \colon \mathcal{X}^d \to \mathcal{X}^d$. Integer discrete flows (IDF) achieve this using additive coupling layers, rounding translation values to the nearest integer and approximating gradients with the straight-through estimator \cite{Hoogeboom2019IntegerDiscreteFlows}; discrete flows \cite{Tran2019DiscreteFlowsInvertible} apply affine coupling layers in modulo space while also restricting the translation and scaling coefficients to a finite number of possible values. In this case the change of variables rule (Eqn. \ref{eqn:flow-change-variables}) simplifies to \cite{Hoogeboom2019IntegerDiscreteFlows, Tran2019DiscreteFlowsInvertible}
\begin{equation}
    p(\vec{x}) = p(f(\vec{x})).
\end{equation}
Unlike the continuous case, there is no Jacobian determinant term; intuitively this term adjusts for volume changes, however, in a discrete space there is no volume. As such, there is no requirement for $f$ to have an efficiently computable Jacobian determinant \cite{Tran2019DiscreteFlowsInvertible}. The absence of this term is restricting, however, discrete flows can only permute the values of $p(\vec{x})$, not change them i.e. a uniform base distribution can only be mapped to another uniform distribution \cite{Papamakarios2019NormalizingFlowsProbabilistic}. Nevertheless, this can be avoided by embedding the data into a space with more values than the data, making IDFs more flexible than discrete flows \cite{berg2020idf++}.

\subsection{Continuous Time Flows}
It is possible to consider a normalizing flow with an infinite number of steps that is defined instead by an ordinary differential equation specified by a Lipschitz continuous neural network $f$ with parameters $\theta$, that describes the transformation of a hidden state $\vec{x}(t) \in \mathbb{R}^D$ \cite{Chen2018NeuralOrdinaryDifferential},
\begin{equation}
    \frac{\partial\vec{x}(t)}{\partial t} = f(\vec{x}(t), t, \theta).
\end{equation}
Starting from input noise $\vec{x}(t_0)$, an ODE solver can solve an initial value problem for some time $t_1$, at which data is defined, $\vec{x}(t_1)$. Modelling a transformation in this form has a number of advantages such as inherent invertibility by running the ODE solver backwards, parameter efficiency, and adaptive computation. However, it is not immediately clear how to train such a model through backpropagation. While it is possible to backpropagate directly through an ODE solver, this limits the choice of solvers to differentiable ones as well as requiring large amounts of memory. Instead, the authors apply the adjoint sensitivity method which instead solves a second, augmented ODE backwards in time and allows the use of a black box ODE solver. That is, to optimise a loss dependent on an ODE solver: 
\begin{equation}
\begin{split}
    \mathcal{L}(\vec{x}(t_1)) &= \mathcal{L}\bigg( \vec{x}(t_0) + \int_{t_0}^{t_1} f(\vec{z}(t),t,\theta)dt \bigg), \\
    &= \mathcal{L}(\text{ODESolve}(\vec{x}(t_0), f, t_0, t_1, \theta)),
\end{split}
\end{equation}
the adjoint $\vec{a}(t)=\frac{\partial \mathcal{L}}{\partial \vec{x}(t)}$ can be used to calculate the derivative of loss with respect to the parameters in the form of another initial value problem \cite{Pontryagin2018MathematicalTheoryOptimal},
\begin{equation}
    \frac{\partial\mathcal{L}}{\partial\theta} = \int_{t_1}^{t_0} \Big(\frac{\partial \mathcal{L}}{\partial \vec{x}(t)}\Big)^T \frac{\partial f(\vec{x}(t), t, \theta)}{\partial \theta} dt,
\end{equation}
which can be efficiently evaluated by automatic differentiation at a time cost similar to evaluating $f$ itself.

Despite the complexity of this transformation, the continuous change of variables rule is remarkably simple:
\begin{equation}
    \frac{\partial \ln p(\vec{x}(t))}{\partial t} = - tr \bigg( \frac{\partial}{\partial\vec{x}(t)} f(\vec{x}(t),t,\theta) \bigg),
\end{equation}
and can be computed using an ODE solver as well. The resulting continuous-time flow is known as FFJORD \cite{Grathwohl2018FFJORDFreeformContinuous}. Since the length of the flow tends to infinity (an infinitesimal flow), the true posterior distribution can be recovered \cite{Rezende2015VariationalInferenceNormalizing}.

As previously mentioned, invertible functions suffer from topological problems; this is especially true for Neural ODEs since their continuous nature prevents trajectories from crossing. Similar to augmented normalizing flows \cite{huang2020augmented}, this can be solved by providing additional dimensions for the flow to traverse \cite{Dupont2019AugmentedNeuralODEs}. Specifically, a $p$-dimensional Euclidean space can be approximated by a Neural ODE in a $(2p+1)$-dimensional space \cite{Zhang2019ApproximationCapabilitiesNeural}. 

\subsubsection{Regularising Trajectories}
ODE solvers can require large numbers of network evaluations, notably when the ODE is stiff or the dynamics change quickly in time. By introducing regularisation, a simpler ODE can be learned, reducing the number of evaluations required. Specifically, all works here are inspired by optimal transport theory to encourage straight trajectories. Monge-Amp\`ere Flow \cite{zhang2018mongeamp} and Potential Flow Generators \cite{yang2020potential} parameterise a potential function satisfying the Monge-Amp\`ere equation \cite{caffarelli1999monge, villani2003topics} with a neural network. RNODE \cite{finlay2020how} applies transport costs to FFJORD as well as regularising the Frobenius norm of the Jacobian, encouraging straight trajectories. By combining these approaches, OT-Flow \cite{Onken2020OTFlowFastAccurate} utilises the optimal transport derivation to derive an exact trace definition with cost similar to stochastic estimators.

\section{Evaluation Metrics}
A huge problem when developing generative models is how to effectively evaluate and compare them. Qualitative comparison of random samples plays a large role in the majority of state-of-the-art works, however, it is subjective and time-consuming to compare many works. Calculating the log-likelihood on a separate validation set is popular for tractable likelihood models but comparison with implicit likelihood models is difficult and while it is a good measure of diversity, it does not correlate well with quality \cite{Theis2016NoteEvaluationGenerative}. 

One approach to quantify sample quality is Inception Score (IS) \cite{Salimans2016ImprovedTechniquesTraining} which takes a trained classifier and determines whether a sample has low label entropy, indicating that a meaningful class is likely, and whether the distribution of classes over a large number of samples has high entropy, indicating that a diverse range of images can be sampled. A perfect IS can be scored by a model that creates only one image per class \cite{Lucic2018AreGANsCreated} leading to the creation of Fr\'echet Inception Distance (FID) \cite{Heusel2017GANsTrainedTwo} which models the activations of a particular layer of a classifier as multivariate Gaussians for real and generated data, measuring the Fr\'echet distance between the two.

These approaches are trivially solved by memorising the dataset and are less applicable to non-natural image-related data. Kernel Inception Distance (KID) \cite{binkowski2018demystifying} instead calculates the squared maximum mean discrepancy in feature space, however, pretrained features may not be sufficient to detect overfitting. Another approach is to train a neural network to distinguish between real and generated samples similar to the discriminator from a GAN; while this detects overfitting, it increases the complexity and time required to evaluate a model and is biased towards adversarial models \cite{gulrajani2019gan}.

\section{Applications}
In general, the definition of a generative model means that any technique can be used on any modality/task, however, some models are more suited for certain tasks. Standard autoregressive networks are popular for text/audio generation \cite{Child2019GeneratingLongSequences, Brown2020LanguageModelsAre, Oord2016WaveNetGenerativeModel}; VAEs have been applied but posterior collapse is difficult to mitigate \cite{bowman2016generating, babaeizadeh2018stochastic}; GANs are more parameter efficient but struggle to model discrete data \cite{nie2019relgan} and suffer from mode collapse \cite{kumar2019melgan}; some normalizing flows offer parallel synthesis, providing substantial speedup \cite{Ziegler2019LatentNormalizingFlows, Tran2019DiscreteFlowsInvertible, prenger2019waveglow}. Video synthesis is more challenging due its exceptionally high dimensionality, typically approaches combine a latent-based implicit generative model to generate individual frames, with an autoregressive network used to predict future latents \cite{lee2018stochastic, babaeizadeh2018stochastic, Kumar2019VideoFlowFlowBasedGenerative} similar to how world models are constructed in reinforcement learning \cite{ha2018world, hafner2019dream}. Modality conversion has been achieved using GANs \cite{Zhu2017UnpairedImageToImageTranslation}, VAE-GANs \cite{liu2017unsupervised}, and DDPMs \cite{sasaki2021unit}.

\subsection{Implicit Representation}
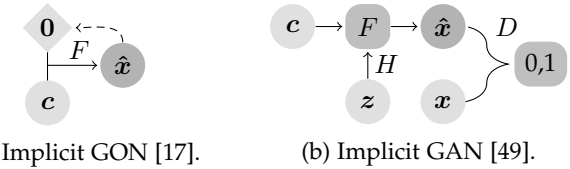
\begin{figure}[t]
\centering
\begin{subfigure}{0.49\linewidth}
    \centering
    \begin{tikzpicture}[shorten >=1pt,->]
        \tikzstyle{cir}=[circle,fill=black!30,minimum size=17pt,inner sep=0pt]
        \tikzstyle{dia}=[diamond,fill=black!30,minimum size=19pt,inner sep=0pt]
        \tikzstyle{box}=[rounded corners=6pt,fill=black!25,minimum size=17pt,inner xsep=4pt, inner ysep=0pt]
        \tikzstyle{sqr}=[fill=black!25,minimum size=17pt, inner ysep=0pt, inner xsep=4pt]
        
        % draw GON
        \draw (0,1) -- (0,0);
        
        \node[dia,fill=black!12]                (GON-0)    at (0,1) {$\vec{0}$};
        \node[cir,fill=black!12]  (GON-1)    at (0,0) {$\vec{c}$};
        \node[cir]                (GON-2)    at (1,0.5) {$\vec{\hat{x}}$};
        \draw (0,0.5) -> (GON-2);
        \draw[densely dashed] (GON-2.north) to [out=90,in=0] (GON-0.east);
        \node[align=center] at (0.42,0.7) {$F$};
        % \node[align=center] at (0.42,1.3) {$\nabla$};
        
    \end{tikzpicture}
    \caption{Implicit GON \cite{Bond-Taylor2020GradientOriginNetworks}.}
    \label{fig:implicit-gon}
\end{subfigure}
\begin{subfigure}{0.49\linewidth}
    \centering
    \begin{tikzpicture}[shorten >=1pt,->]
        \tikzstyle{cir}=[circle,fill=black!30,minimum size=17pt,inner sep=0pt]
        \tikzstyle{dia}=[diamond,fill=black!30,minimum size=19pt,inner sep=0pt]
        \tikzstyle{box}=[rounded corners=6pt,fill=black!25,minimum size=17pt,inner xsep=4pt, inner ysep=0pt]
        \tikzstyle{sqr}=[fill=black!25,minimum size=17pt, inner ysep=0pt, inner xsep=4pt]
        
        \node[cir]                (GAN-xh)   at (2,1) {$\vec{\hat{x}}$};
        \node[box]                (GAN-F)    at (1,1) {$F$};
        \node[cir,fill=black!12]  (GAN-z)    at (1,0) {$\vec{z}$};
        \node[cir,fill=black!12]  (GAN-x)    at (2,0) {$\vec{x}$};
        \node[cir,fill=black!12]  (GAN-c)    at (0,1) {$\vec{c}$};
        \node[box]                (GAN-rf)   at (3.3,0.5) {0,1};
        \draw[>=, decorate, decoration={brace, amplitude=16pt, mirror}] (2.3,0) -- coordinate [left=10pt] (B) (2.3,1) node {};
        
        \draw(GAN-z) -- (GAN-F);
        \draw(GAN-F) -- (GAN-xh);
        \draw(GAN-c) -- (GAN-F);

        \node[align=center] at (2.85,1) {$D$};
        \node[align=center] at (1.25,0.5) {$H$};
    \end{tikzpicture}
    \caption{Implicit GAN \cite{dupont2021generative}.}
    \label{fig:implicit-gan}
\end{subfigure}
\caption{Implicit networks model data continuously permitting arbitrarily high resolutions. Dashed lines represent gradients, $F$ is an implicit network, and $H$ is a hypernetwork.}
\label{fig:implicit-networks}
\end{figure}

Typically deep architectures discussed in this survey are built with data represented as discrete arrays thus using discrete components such as convolutions and self-attention. Implicit representation on the other hand treats data as continuous signals, mapping coordinates to data values \cite{Sitzmann2020ImplicitNeuralRepresentationsa, tancik2020fourier}. Implicit Gradient Origin Networks (GONs; Fig. \ref{fig:implicit-gon}) \cite{Bond-Taylor2020GradientOriginNetworks} form a latent variable model by concatenating latent vectors with coordinates which are passed through an implicit network; here latent vectors are calculated as the gradient of a reconstruction loss with respect to the origin. By sampling using a finer grid of coordinates, super-resolution beyond resolutions seen during training is possible. Other approaches to learn an implicit generative model as a GAN include directly feeding latents through an implicit network with upsampling \cite{karras2021aliasfree} and mapping latents to the weights of an implicit function using a hyper-network \cite{dupont2021generative} (Fig. \ref{fig:implicit-gan}).

\section{Conclusion}
While GANs have led the way in terms of sample quality for some time now, the gap between other approaches is shrinking; the diminished mode collapse and simpler training objectives make these models more enticing than ever, however, the number of parameters required in addition to slow run-times pose a substantial handicap. Despite this, recent work in hybrid models offers a balance between extremes at the expense of extra model complexity that hinders broader adoption. The varied connections between these systems mean that advances in one field inevitably benefit others, for instance, improved variational bounds are beneficial for VAEs, diffusion models, and surjective flows, and the application of innovative data augmentation strategies has been found to offer benefits across numerous model classes without necessitating more powerful architectures. When it comes to scaling models to high-dimensional data, attention is a common theme, allowing long-range dependencies to be learned; recent advances in linear attention will aid scaling to even higher resolutions. Implicit networks are another promising direction, allowing efficient synthesis of arbitrarily high resolution and irregular data. Similar unified generative models capable of modelling continuous, irregular, and arbitrary length data, over different scales and domains will be key for the future of generalisation.

\ifCLASSOPTIONcaptionsoff
  \newpage
\fi

\bibliographystyle{plain}
\bibliography{references_abbrev}

\vfill

\vspace{-2.8em}
\vspace{-5.7em}
\begin{IEEEbiography}[{\vspace*{-1em}\includegraphics[width=1in,clip,keepaspectratio]{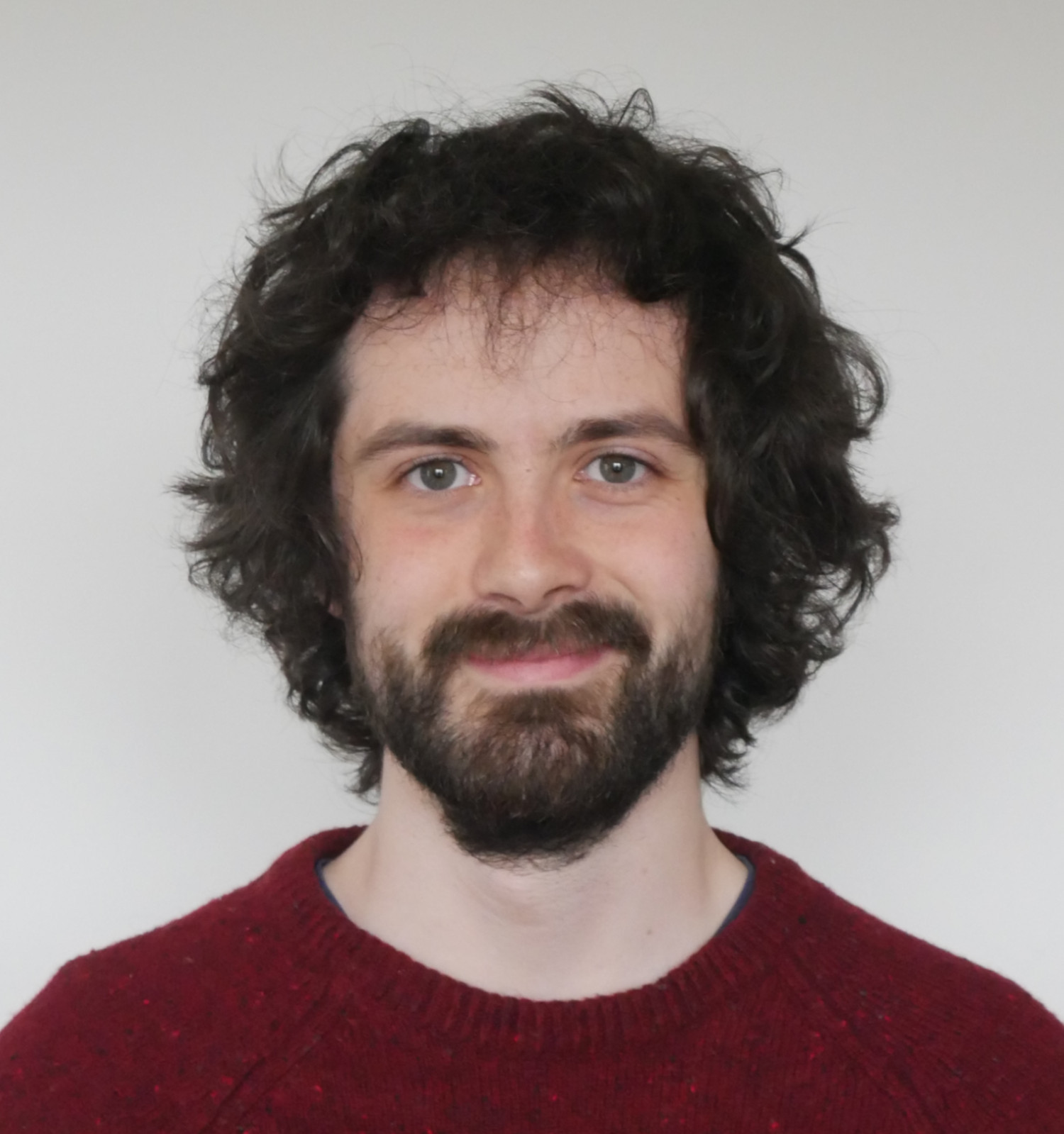}}]{Sam Bond-Taylor}
is a PhD student in the Department of Computer Science at Durham University. His research interests are focused around unsupervised deep learning methods and the connections with human learning. In particular, the development of generative models and machine reasoning systems. He is a teaching assistant on the deep learning module at Durham University.
\end{IEEEbiography}

\vspace{-4.5em}
\begin{IEEEbiography}[{\vspace*{-2em}\includegraphics[width=1in,clip,keepaspectratio]{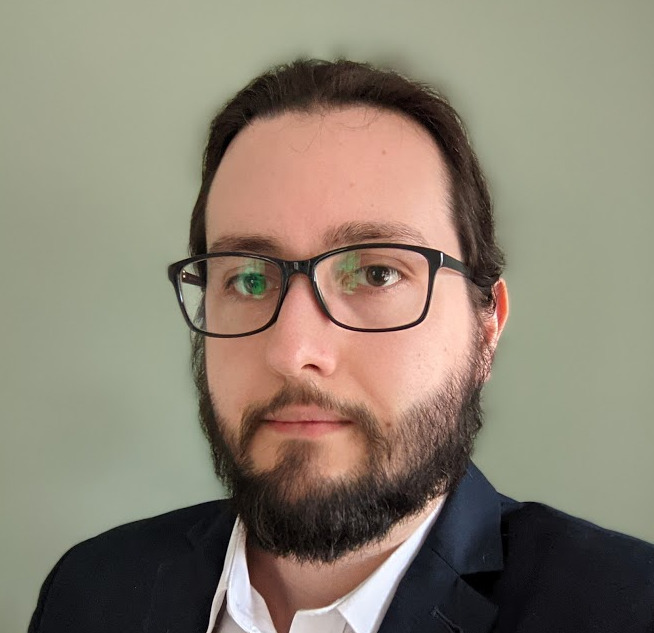}}]{Adam Leach}
is a PhD student in the Department of Computer Science at Durham University. His research focuses on  applying deep generative models and reinforcement learning techniques to molecular modelling problems such as protein folding and docking. He is a teaching assistant on the reinforcement learning module at Durham University.
\end{IEEEbiography}

\vspace{-5.3em}
\begin{IEEEbiography}[{\includegraphics[width=1in,height=1.25in,clip,keepaspectratio]{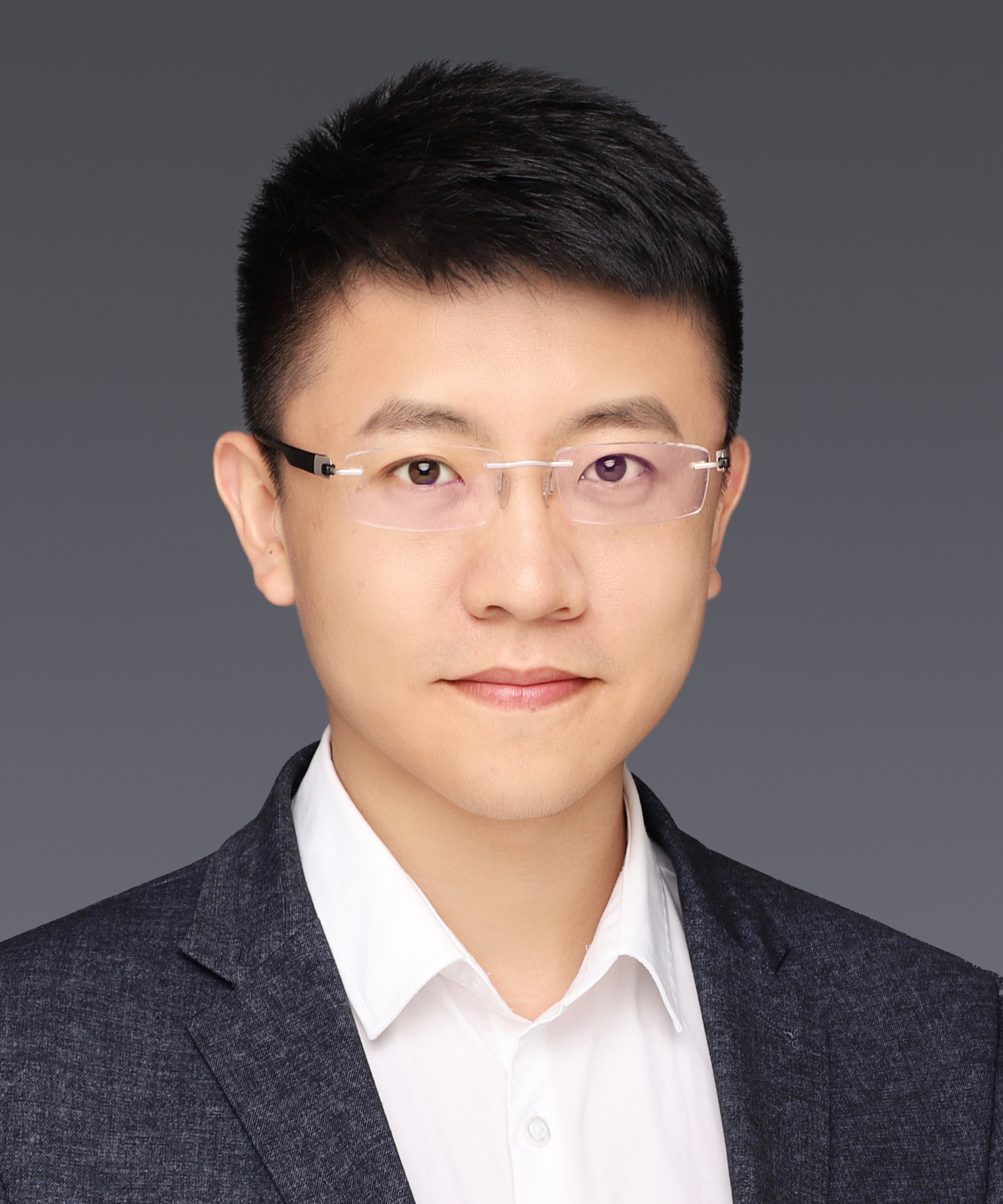}}]{Yang Long}
is an assistant professor in the department of computer science, Durham University. He is also an MRC innovation fellow aiming to design scalable AI solutions for large-scale healthcare applications. His research background is in the highly interdisciplinary field of computer vision and machine learning. 
% While he is passionate about unveiling the black-box of AI brain and transferring the knowledge to seek Scalable, Interactable, Interpretable, and sustainable solutions for other disciplinary researches, e.g. physical activity, mental health, design, education, security, and geoengineering. 
He has authored/co-authored 20+ top-tier papers in refereed journals/conferences such as IEEE TPAMI, TIP, CVPR, AAAI, and ACM MM.
\end{IEEEbiography}

\vspace{-4.5em}
\begin{IEEEbiography}[{\vspace*{-1em}\includegraphics[width=1in,height=1.25in,clip,keepaspectratio]{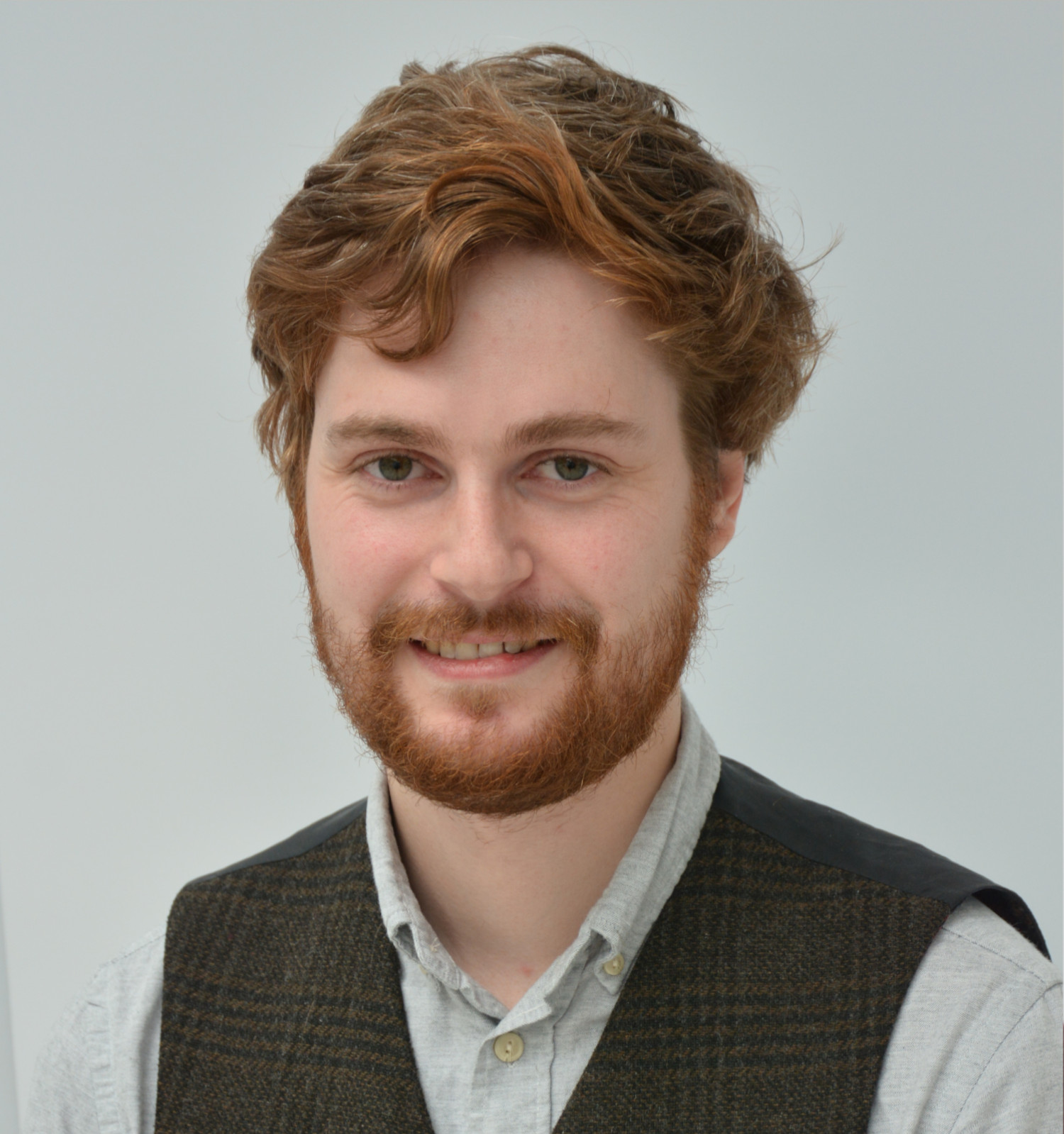}}]{Chris G. Willcocks}
is an assistant professor in computer science at Durham University, where his interdisciplinary research focuses on generative models, medical image computing, computational biophysics and machine reasoning. He teaches deep learning, reinforcement learning and cyber security, and publishes top-tier journal and conference papers in venues such as ICLR, PRX, IEEE TPAMI, TMI and TIFS.
\end{IEEEbiography}

\end{document}